%% file: main.tex
\documentclass[pdflatex,sn-mathphys-num]{sn-jnl}

\usepackage{graphicx}%
\usepackage{multirow}%
\usepackage{amsmath,amssymb,amsfonts}%
\usepackage{amsthm}%
\usepackage{mathrsfs}%
\usepackage[title]{appendix}%
\usepackage{xcolor}%
\usepackage{textcomp}%
\usepackage{manyfoot}%
\usepackage{makecell}
\usepackage{booktabs}%
\usepackage{algorithm}%
\usepackage{algorithmicx}%
\usepackage{algpseudocode}%
\usepackage{listings}%
\usepackage{url}%
\usepackage{tabularx}%
\usepackage{adjustbox}%
\usepackage[justification=centering]{caption}%
\usepackage{tikz}%
\usepackage{longtable}%
\usepackage[T1]{fontenc}
\usepackage{lmodern}
\usepackage{amsmath,amssymb}
\usepackage{hyperref}
\usepackage{xcolor}
\usepackage{listings}
\usetikzlibrary{shapes.geometric, arrows.meta, positioning}
\usepackage{graphicx}
\usepackage{caption}
\usepackage{subcaption}
\usepackage{tikz}
\theoremstyle{thmstyleone}

\theoremstyle{thmstyletwo}%
\theoremstyle{thmstylethree}%

\usepackage[most]{tcolorbox}
\tcbuselibrary{breakable, skins, theorems}
\usepackage{pifont}
\usepackage{booktabs}
\usepackage{graphicx}

\tcbset{
  colback=white,
  colframe=black!60,
  coltitle=black,
  colbacktitle=black!10,
  fonttitle=\bfseries,
  boxrule=0.5pt,
  arc=3mm,
  outer arc=3mm,
  sharp corners=south,
  enhanced,
  breakable,
  before upper={\parindent15pt},
}

\begin{document}

\title[Article Title]{A Graph-Retrieval-Augmented Generation Framework Enhances Decision-Making in the Circular Economy}

\author[1]{\fnm{Yang} \sur{Zhao}}\email{Zhao$\_$Yang@simtech.a-star.edu.sg}

\author[2]{\fnm{Chengxiao} \sur{Dai}}\email{daicxx1226@gmail.com}

\author[3]{\fnm{Dusit} \sur{Niyato}}\email{DNIYATO@ntu.edu.sg}

\author[1]{\fnm{Chuan Fu} \sur{Tan}}\email{Tan$\_$Chuan$\_$Fu@simtech.a-star.edu.sg}

\author[3]{\fnm{Keyi} \sur{Xiang}}\email{KEYI003@e.ntu.edu.sg}

\author[4]{\fnm{Yueyang} \sur{Wang}}\email{e0726339@u.nus.edu}

\author[1]{\fnm{Zhiquan} \sur{Yeo}}\email{zqyeo@simtech.a-star.edu.sg}

\author[1]{\fnm{Daren} \sur{Tan Zong Loong}}\email{tanzld@simtech.a-star.edu.sg}

\author[1]{\fnm{Jonathan} \sur{Low Zhaozhi}}\email{Jonathan$\_$Low@simtech.a-star.edu.sg}

\author[1]{\fnm{Eugene} \sur{H.Z. HO}}\email{eugene$\_$ho@simtech.a-star.edu.sg}

\affil[1]{\orgdiv{Singapore Institute of Manufacturing Technology (SIMTech)}, \orgname{Agency for Science, Technology and Research (A*STAR)}, \orgaddress{\street{2 Fusionopolis Way, Innovis
$\#$08-04}, \postcode{138634}}, \country{Republic of Singapore}}

\affil[2]{\orgdiv{School of Computer Science, Faculty of Engineering}, \orgname{The University of Sydney}, \orgaddress{\street{Building J12, 1 Cleveland Street, Darlington}, \state{NSW}, \postcode{2008}}, 
\country{Australia}}

\affil[3]{\orgdiv{College of Computing and Data Science}, \orgname{Nanyang Technological University}, \orgaddress{\street{50 Nanyang Ave}, \postcode{639798}}, \country{Republic of Singapore}}

\affil[4]{\orgdiv{Faculty of Science}, \orgname{National University of Singapore}, \orgaddress{\street{6 Science Drive 2}, \postcode{117546}}, \country{Republic of Singapore}}


\abstract{
Large language models (LLMs) hold promise for sustainable manufacturing, but often hallucinate industrial codes and emission factors, undermining regulatory and investment decisions. We introduce CircuGraphRAG, a retrieval-augmented generation (RAG) framework that grounds LLMs outputs in a domain-specific knowledge graph for the circular economy. This graph connects 117,380 industrial and waste entities with classification codes and GWP100 emission data, enabling structured multi-hop reasoning. Natural language queries are translated into SPARQL and verified subgraphs are retrieved to ensure accuracy and traceability. Compared with Standalone LLMs and Naive RAG, CircuGraphRAG achieves superior performance in single-hop and multi-hop question answering, with ROUGE-L F1 scores up to 1.0, while baseline scores below 0.08. It also improves efficiency, halving the response time and reducing token usage by 16\% in representative tasks. CircuGraphRAG provides fact-checked, regulatory-ready support for circular economy planning, advancing reliable, low-carbon resource decision making.
}

\keywords{Circular Economy, Large Language Model, RAG, GraphRAG.}

\maketitle

\input{Sections/1-introduction}

\input{Sections/2-results}

\input{Sections/3-discussion}

\input{Sections/4-methods}
\input{Sections/5-conclusion}
\bibliography{sn-bibliography}

\input{Sections/6-appendix}

\end{document}

%% file: Sections/1-introduction.tex
\section{Introduction}\label{sec1}

The global transition to a circular economy promises substantial improvements in resource efficiency and waste reduction, whereby the by-products of one industrial process can serve as input to another, thus reducing waste and conserving resources~\cite{chertow2000industrial,lombardi2012redefining}. Despite its considerable potential, many organizations face persistent barriers, including fragmentation of data, limited trust, and stringent regulatory requirements, which hinder their ability to align resource availability with sustainability objectives~\cite{Puchel2024,su16146061,nookala2024governance}. In addition, material flow, energy use, and emission data are frequently located in isolated systems, undermining collective efforts to minimize waste~\cite{Kullmann2021,inbook}. Even when potential opportunities arise, the lack of standardized terminology, divergent measurement protocols, and intellectual property concerns often restrict open data sharing, thus impeding the widespread adoption of circular practices~\cite{MORAGA2019452,ROY2022113}.

Currently, environmental, social, and governance (ESG) reporting frameworks~\cite{tsang2023environmental}, such as the Global Reporting Initiative (GRI)~\cite{hedberg2003global}, the Sustainability Accounting Standards Board (SASB)~\cite{hales2021sustainability}, and the Climate-related Financial Disclosure Task Force (TCFD)~\cite{board2017task}, have gained prominence in guiding corporate transparency and investment decisions. These frameworks require auditable disclosures on waste streams, resource utilization, and environmental impacts, aligning closely with the objectives of a circular economy. However, bridging data fragmentation, establishing trust across organizations, and meeting compliance demands remain non-trivial tasks.

Standardized databases and systematic monitoring frameworks have emerged to address these data-related challenges~\cite{fraccascia2018role,fraccascia2020and}. Classification systems such as the European Waste Catalog (EWC)~\cite{van2021toward} and the Nomenclature of Economic Activities (NACE)~\cite{vidali2024unlocking} provide universal taxonomies for categorizing resources and waste, thus reducing intersectoral misalignment. These foundational structures are suitable for the construction of knowledge graphs~\cite{hogan2021knowledge}, where entities such as waste streams, production facilities are represented as nodes, and their relationships, including ``can be processed by,''  and ``located in'' are encoded as semantic links. Typically, these links are implemented using the Resource Description Framework (RDF)~\cite{powers2003practical, lassila1997resource}, ensuring structured and machine-readable data across multiple domains.

Recent advances in large language models (LLMs) offer expert-level insight and decision support in various fields, including healthcare and environmental management~\cite{radford2018improving}. However, LLMs can produce inaccurate or fabricated content due to hallucination~\cite{rillig2023risks,zhu2023chatgpt,ji2023survey}, reinforcing the need for robust and vetted databases. Retrieval-augmented generation (RAG)~\cite{lewis2020retrieval,gao2023retrieval} partially addresses this challenge by integrating external knowledge into the inference process. An advanced variant, GraphRAG~\cite{hu2024grag}, exploits the interconnected nature of knowledge graphs to perform multi-hop reasoning, producing more reliable outputs than traditional RAG pipelines based solely on vector similarity~\cite{gao2024retrievalaugmentedgenerationlargelanguage,lewis2021retrievalaugmentedgenerationknowledgeintensivenlp}. By systematically linking nodes through explicit relationships, GraphRAG retrieves contextually relevant knowledge, thus improving precision and mitigating tangential data~\cite{kang2023knowledgegraphaugmentedlanguagemodels}. In addition, this graph-based structure clarifies the chain of reasoning, allowing users to trace how each data point contributes to the final output, mitigating hallucinations, and increasing user trust.

To align with climate mitigation targets, it is also critical to integrate emission factor data. The 100-year Global Warming Potential (GWP100) indicator provides a standardized benchmark to assess the climate impacts of industrial processes~\cite{godal2002testing,epa2023gwp}. Reflecting GWP100 in transactions, stakeholders can rank or filter synergy opportunities based on greenhouse gas footprints, thus making decisions that comply with ESG standards and advance sustainability goals.

Together, these frameworks and technologies create the foundation for a robust, data-driven platform that can effectively enable circular economy synergies. Despite these promising developments, direct applications of knowledge graphs and GraphRAG within the circular economy remain limited. Existing standards and frameworks provide a foundation for modeling waste and resource data but lack integration into a cohesive platform capable of supporting large-scale synergy discovery and continuous sustainability assessments~\cite{han2025retrievalaugmentedgenerationgraphsgraphrag,peng2024graphretrievalaugmentedgenerationsurvey}.

In this paper, we detail how CircuGraphRAG synergistically integrates knowledge graph technology, RAG, and ESG metrics to advance data-driven circular economy practices. Specifically, we build a harmonized industrial-waste knowledge graph that links industries, waste classifications, materials, and environmental factors (GWP100) across multiple standards (ISIC, NACE, EWC), enabling structural and semantic exploration. We then present a knowledge graph enabled RAG architecture that incorporates ontology-aligned industrial knowledge and environmental metrics, supporting structured retrieval and multi-hop reasoning for low-carbon industrial symbiosis. Our approach further employs a light-weight library of 18 SPARQL query templates, automatically selected and parameterized by the LLMs, to facilitate both single-hop lookups and multi-hop synergy discovery without handwritten rules. In addition, a hybrid ranking mechanism uses GWP100 to order query results, guiding practitioners toward low-carbon waste-to-resource pathways. We also contribute a reproducible graph reasoning benchmark for code alignment, numeric accuracy, and industrial symbiosis scenarios. Finally, we demonstrate model-agnostic empirical gains across various LLMs, showing that CircuGraphRAG consistently reduces hallucinations and accelerates inference compared to Naive RAG, without diminishing answer completeness.

\vspace{1em}
\noindent The abbreviations used in the paper are summarized in the Appendix~\ref{sec:Abbreviations}.

%% file: Sections/2-results.tex
\section{Results}
\label{sec:results}

Developing robust waste-to-resource strategies is crucial to minimize environmental impacts and advance circular economy goals. This section presents an extensive evaluation of CircuGraphRAG across realistic waste-to-resource management queries, illustrating how knowledge-graph-driven retrieval substantially enhances the discovery of reuse pathways. We compare our CircuGraphRAG against several baselines on three primary tasks: structured data extraction (single-hop queries) and multi-hop reasoning including synergy identification. Quantitative metrics are complemented by an analysis of computational overhead and template matching ablation study, as well as an assessment of the multi-round consistency of LLMs reasoning.

\subsection{Quantitative Results for Single-Hop Queries}
\label{subsec:single-hop}

Single-hop queries consist of directly retrieving a single fact or entity from the knowledge graph. A representative query is \textit{``Which resource corresponds to EWC code 080121 and HS code 810330?"} in Appendix~\ref{sec:groundtruth-examples}. Then, Appendix~\ref{sec:single-hop-metrics} summarizes the results on three distinct single-hop queries (Cases~1--3).

\subsubsection{Overall Performance}
CircuGraphRAG consistently outperforms both Standalone LLM and Naive RAG across all examined LLMs (Llama, Qwen, DeepSeek) in terms of \textsc{ROUGE-L} F1 scores. These improvements are visualized in \autoref{fig:singlehop-3d}, which illustrates the performance differences across models and test cases in three dimensions. Detailed numerical results can be found in Appendix~\ref{sec:single-hop-metrics}, highlighting the significant gains enabled by domain-constrained retrieval.

\subsubsection{Precision and Recall Gains}
In Case~1, CircuGraphRAG with Qwen and with DeepSeek both achieve a perfect \textsc{ROUGE-L} recall (1.0). In contrast, Standalone LLM approaches frequently produce incomplete or inaccurate responses, suggesting hallucinations or a lack of contextual grounding. However, CircuGraphRAG anchors the query to specific domain edges, such as \texttt{iskg:hasHSCode} and \texttt{iskg:hasEwcCode}, to reliably match the correct entity.

In Case~2, CircuGraphRAG improves \textsc{ROUGE-L} precision and recall over baselines. For example, CircuGraphRAG with Qwen achieves a \textsc{ROUGE-L} precision of 0.4286, a substantial gain over Naive RAG with Qwen (0.0150). These improvements indicate that forcing query resolution through valid code and entity relationships (e.g., \texttt{iskg:hasNaceCode}) filters out spurious completions. From a regulatory point of view, an incorrect assignment of provider or resource codes (for example, NACE 3821 vs. 3822) could lead to non-compliance. 

\subsubsection{Numeric Retrieval}
Numerical accuracy is critical for compliance driven domains, particularly for GWP100. In Case~3, CircuGraphRAG with both Qwen and Llama correctly retrieves the minimal GWP100 value (\(0.008930631\)), achieving an \textsc{ROUGE-L} F1 score of 1.0. All baseline approaches either return a spurious figure or no result, underscoring the difficulty of numeric retrieval within unconstrained LLMs. Since small deviations in emission metrics can misrepresent environmental impact assessments, CircuGraphRAG's exact retrieval of numeric data is vital for reliable industrial decision-making.




\begin{figure}[!ht]
    \centering
    \begin{subfigure}[t]{1.0\textwidth}
        \centering
        \hspace*{-2.35cm}
        \includegraphics[width=1.28\textwidth]{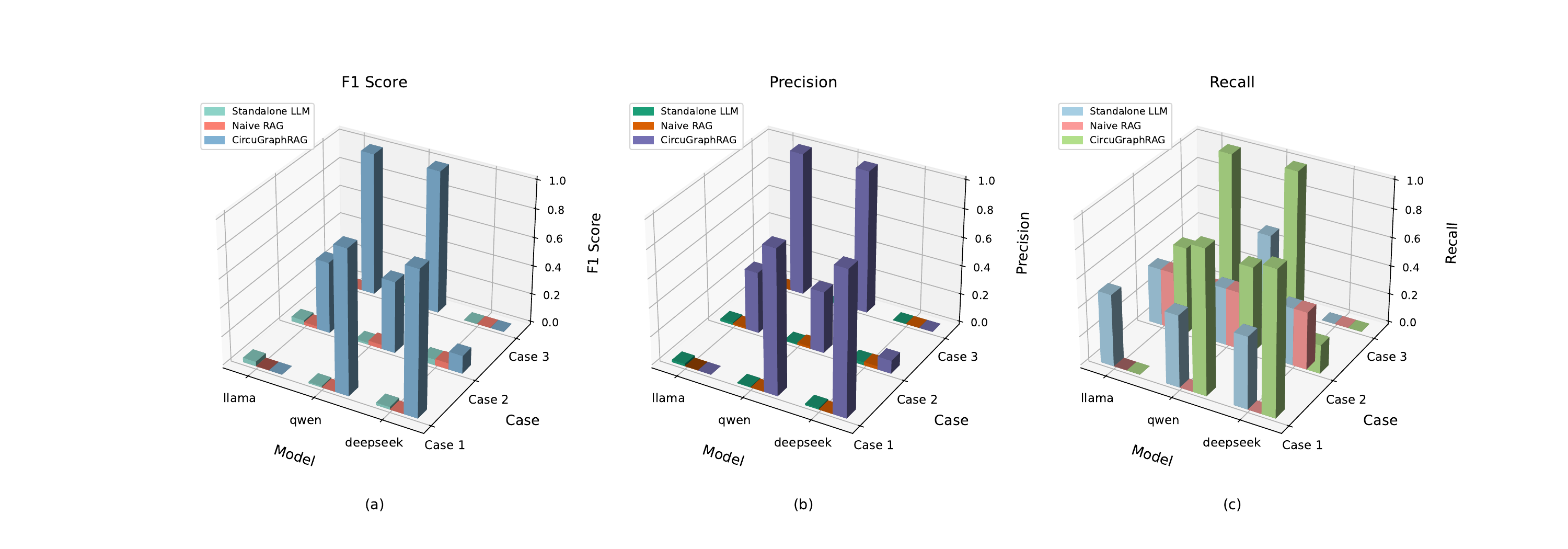}
        \caption{Single-hop QA: ROUGE-L F1, precision, and recall across Cases 1--3.}
        \label{fig:singlehop-3d}
    \end{subfigure}
    
    \vspace{0.4cm} 
    
    \begin{subfigure}[t]{1.0\textwidth}
        \centering
        \hspace*{-2.35cm}
        \includegraphics[width=1.28\textwidth]{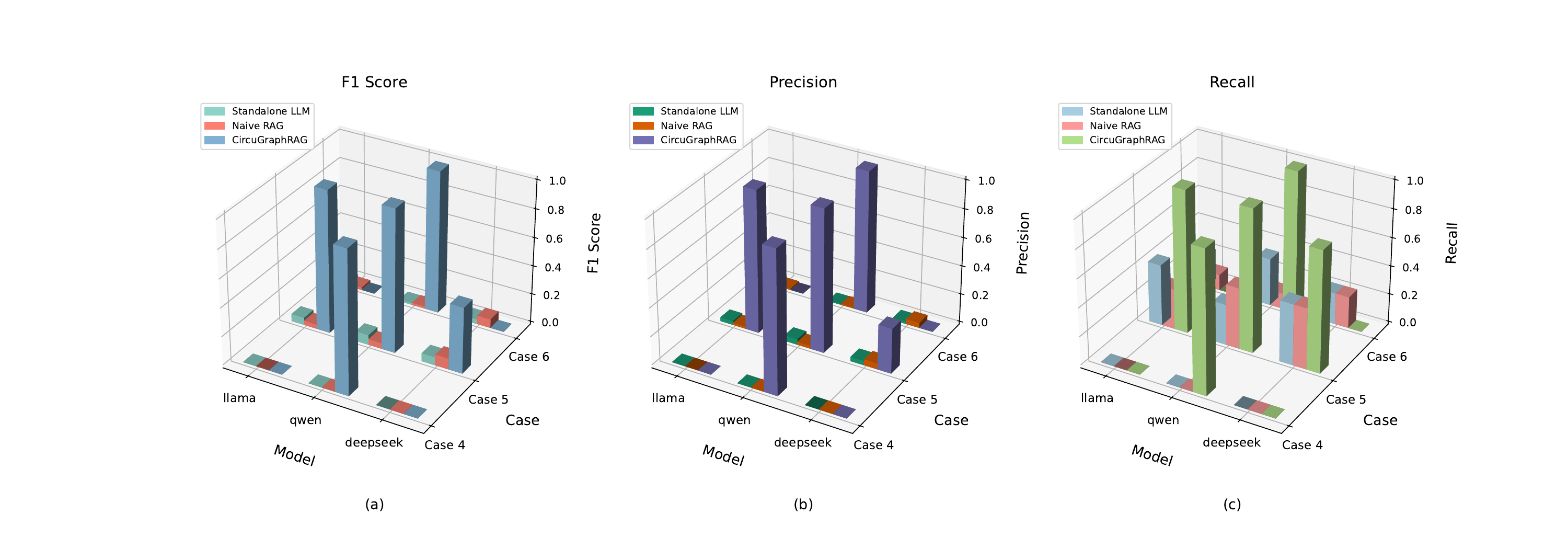}
        \caption{Multi-hop QA: ROUGE-L F1, precision, and recall across Cases 4--6.}
        \label{fig:multihop-3d}
    \end{subfigure}
    
    \caption{3D performance comparison of different LLMs and retrieval methods on single-hop QA tasks.}
    \label{fig:combined-qa-3d}
\end{figure}

\subsection{Quantitative Results for Multi-Hop Reasoning and Synergy Identification}
\label{subsec:multi-hop}

Multi-hop queries assess a model’s capacity to traverse multiple linked relationships, an essential capability for identifying synergies in industrial symbiosis. For instance, verifying that a receiver of one resource can act as a provider of another often requires navigation through various classification codes (e.g., EWC, HS, NACE, CPA) and properties (e.g.,\texttt{ iskg:hasGwp100}). As shown in \autoref{fig:multihop-3d}, CircuGraphRAG consistently outperforms both Standalone LLM and Naive RAG in Cases~4--6, achieving notably higher \textsc{ROUGE-L}~F1 scores.

In Case~4, the query determines which CPA code and category apply to the resources generated under a specific NACE code. Both Standalone LLM and Naive RAG yield a \textsc{ROUGE-L}~F1 of 0.0 across all language models. CircuGraphRAG with Llama or DeepSeek also returns 0.0, yet CircuGraphRAG with Qwen achieves 1.0, indicating that a sufficiently capable model can take advantage of the constraints based on the knowledge graph.  

Moreover, Case~5 extends this concept by examining whether receivers under a CPA code can later serve as providers of a second resource (e.g. ``waste polystyrene''). CircuGraphRAG with Llama or Qwen achieves a perfect \textsc{ROUGE-L}~F1 of 1.0, whereas Naive RAG and Standalone LLM each fall below 0.08 in most configurations, underscoring that purely generative reasoning is prone to errors without explicit domain guidance. CircuGraphRAG with DeepSeek performs moderately well (\textsc{ROUGE-L}~F1=0.4615), though not matching the near-ideal scores of  CircuGraphRAG with Qwen or Llama.

In Case~6, the query identifies providers of a resource coded EWC~070213 that can also receive ``aluminum scrap.'' CircuGraphRAG with Qwen again achieves a \textsc{ROUGE-L}~F1 of 1.0, while CircuGraphRAG with Llama and DeepSeek remain at 0.0, and the baseline approaches exhibit limited precision below 0.035. These results demonstrate that CircuGraphRAG significantly outperforms baseline methods in complex multi-hop queries, particularly when structural constraints are applied. However, performance differences across LLMs remain, and some models still require strategy calibration or further fine-tuning to fully leverage knowledge graph-enhanced reasoning.

From a circular economy perspective, this level of multi-hop precision is crucial for industrial symbiosis, in which by-products from one facility become feedstock for another. By encoding explicit relationships in a knowledge graph, CircuGraphRAG reduces reliance on heuristic inference, thus mitigating misclassification risks and facilitating data-driven synergy discovery. Overall, these results confirm that CircuGraphRAG enhances multi-hop reasoning, particularly when deployed with capable LLMs, thus improving the identification of multi-step resource flows and advancing circular economy's objectives. Supporting metrics for this evaluation are presented in Appendix~\ref{sec:multi-hop-metrics}.

\subsection{Computation Cost and Token Usage}
\label{subsec:cost-token}

We further evaluated CircuGraphRAG in terms of token consumption and inference time across various LLM back-ends and task types. Although the framework incorporates additional retrieval components including template matching (TM), query merging (QM), and querying with recovered context (QRC), which might increase computational overhead, our results indicate that CircuGraphRAG consistently reduces both total token usage and inference latency. These gains are attributed to the construction of more compact, domain-specific input contexts that constrain the language model’s generative scope and streamline the reasoning process.

\begin{figure}[!ht]
    \centering
    \begin{subfigure}[t]{\textwidth}
        \centering
        \includegraphics[width=\textwidth]{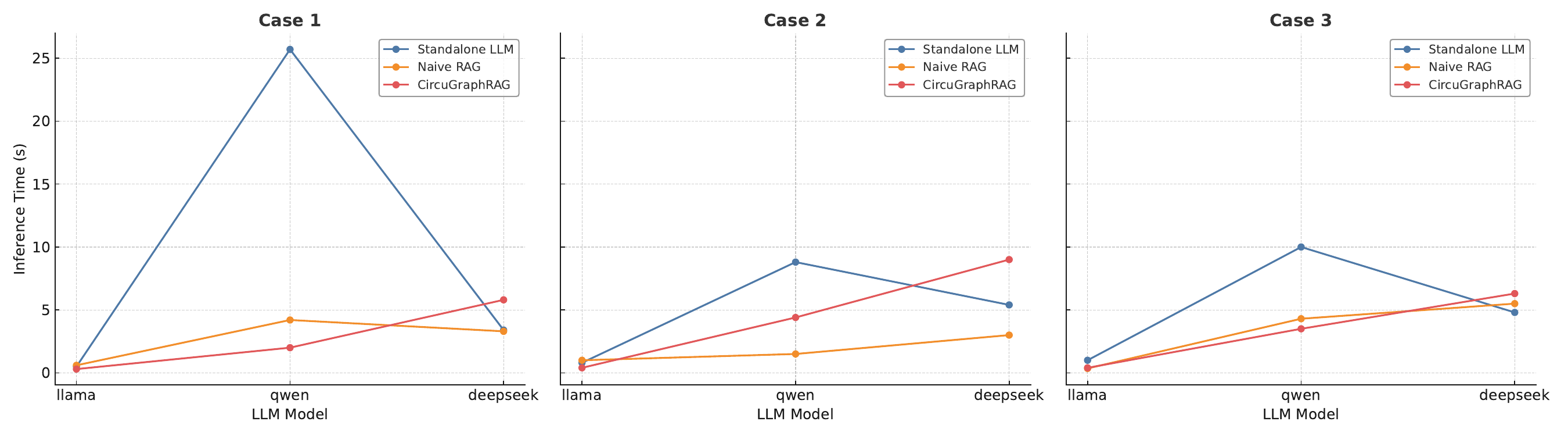}
        \caption{Single-hop cases (Cases 1--3).}
        \label{fig:singlehop-inference}
    \end{subfigure}
    
    \vspace{0.4cm} 

    \begin{subfigure}[t]{\textwidth}
        \centering
        \includegraphics[width=\textwidth]{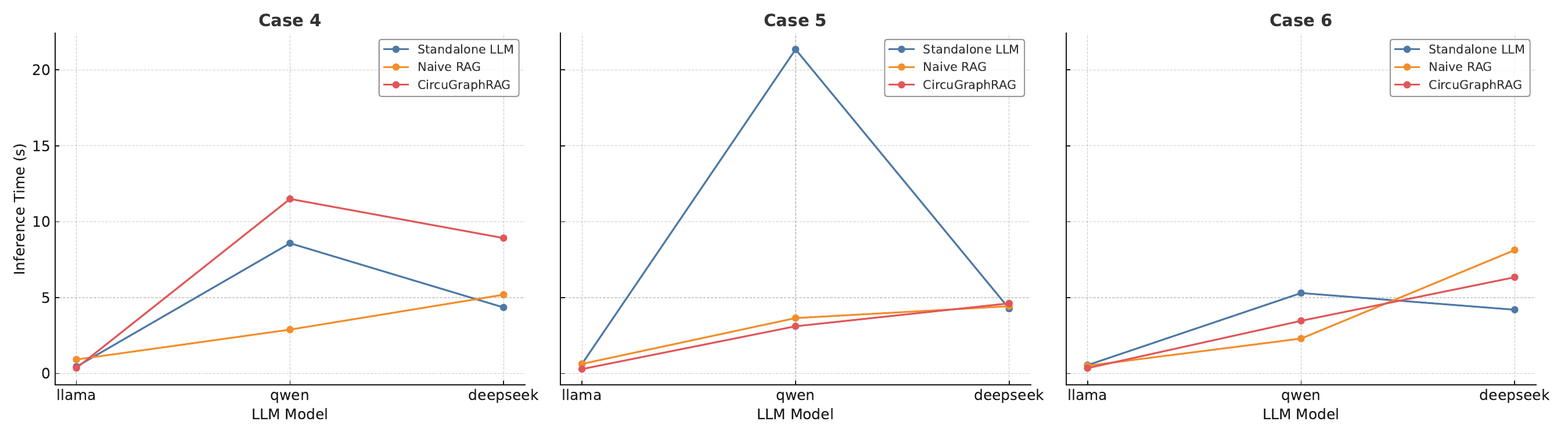}
        \caption{Multi-hop cases (Cases 4--6).}
        \label{fig:multihop-inference}
    \end{subfigure}
    
    \caption{Inference time comparison across different models and methods for (a) single-hop and (b) multi-hop QA tasks.}
    \label{fig:combined-inference}
\end{figure}

In single-hop queries (\autoref{fig:singlehop-inference}), CircuGraphRAG with Qwen processes fewer tokens more quickly than Naive RAG with Qwen or Standalone Qwen, as evidenced in Case~1 (2,009 tokens vs.\ 2,381 tokens and over 10,000 tokens) while doubling inference speed. The approach may incur higher token usage in more complex tasks (Case~2) due to broader subgraph lookups, but yields more reliable answers by leveraging domain-specific evidence. In contrast, Standalone Qwen tends to generate excessively long outputs with correspondingly higher latency.

Multi-hop queries often exacerbate off-topic content unless retrieval is carefully guided. In Case~5 (\autoref{fig:multihop-inference}), Standalone Qwen outputs 8,677 tokens in 21.35\,s, while CircuGraphRAG with Qwen trims responses to 2,733 tokens only in 3.11\,s. Although certain queries (e.g., Case~4) require more extensive retrieval (6,041 tokens over 11.50\,s), CircuGraphRAG is the only method that achieves an F1 score of 1.0 for this scenario. Across all LLMs, CircuGraphRAG constrains generative pathways by embedding domain relationships, balancing overhead with enhanced multi-hop accuracy.

Overall, these observations show that CircuGraphRAG effectively reduces superfluous text generation while sustaining high factual fidelity. When additional tokens are needed, the resulting improvements in answer quality and reliability offset the higher computational cost, demonstrating the potential of a domain-targeted retrieval structure for large-scale industrial knowledge queries.A detailed comparison of token usage and inference latency across models and methods supporting these findings is presented in Appendix~\ref{sec:singlehop-times} and Appendix~\ref{sec:multihop-times}.

\subsection{Template Matching Ablation Study}
Since structural templates play a central role in shaping how the model identifies and composes query paths, we conduct an ablation study to investigate how different levels of template guidance affect the reasoning process in CircuGraphRAG. Specifically, our objective is to evaluate whether removing or weakening the use of templates impacts the accuracy of downstream SPARQL query generation and final answers.

We consider three template configurations: (1) \textbf{with template}, where the system utilizes explicitly defined query structures to guide the reasoning process; (2) \textbf{no template}, where the LLM must generate the SPARQL query path without structural guidance; and (3) \textbf{fuzzy template}, where only partial structural hints (e.g. entity types or field names) are provided without complete logical templates. For each setting, we evaluate the accuracy of the answer in three LLMs under the same task input. To isolate the influence of template guidance, we conducted this study in a representative single-hop QA case where the expected output depends on precise entity-code mapping.

\begin{table}[!ht]
\centering
\caption{Accuracy Comparison under Different Template Matching Conditions. (A \textcolor{blue}{\ding{51}} indicates a correct answer; a \textcolor{red}{\ding{55}} indicates an incorrect one.)}
\label{tab:template-ablation}
\begin{tabular}{lccc}
\toprule
\textbf{Template Type} & \textbf{llama} & \textbf{qwen} & \textbf{deepseek} \\
\midrule
With Template  & \textcolor{blue}{\ding{51}} & \textcolor{blue}{\ding{51}} & \textcolor{blue}{\ding{51}} \\
No Template    & \textcolor{red}{\ding{55}}  & \textcolor{red}{\ding{55}}  & \textcolor{red}{\ding{55}} \\
Fuzzy Template & \textcolor{blue}{\ding{51}} & \textcolor{blue}{\ding{51}} & \textcolor{red}{\ding{55}} \\
\bottomrule
\end{tabular}
\end{table}

As shown in \autoref{tab:template-ablation}, the performance of different models varies significantly across template configurations, further validating the critical role of predefined templates in the CircuGraphRAG system. Under the \textbf{with template} condition, all three models achieve 100\% query matching accuracy. This demonstrates that explicitly injected templates not only provide structural guidance, but also normalize the reasoning path and entity operation process, directly determining the quality and stability of SPARQL query generation.

In contrast, when the templates are completely removed (\textbf{no template}), none of the models can generate correct SPARQL queries, and the accuracy drops to 0. This result clearly demonstrates that even in single-hop question answering scenarios, large language models struggle to reconstruct the necessary triple patterns and query logic without explicit structural templates. The absence of templates leads to significant errors in entity selection, property targeting, and predicate direction, often resulting in failed path construction or invalid answers.

Under the \textbf{fuzzy template} condition, the system provides only partial structural cues. Llama and Qwen still maintain relatively high accuracy, whereas DeepSeek shows a notable performance drop. This suggests that while fuzzy template hints can assist in constructing basic reasoning structures. However, without complete template logic, query path generation becomes unstable and the final answers less reliable, particularly for models with limited structural understanding.

\subsection{Multi-Round Consistency of LLMs Reasoning}

To evaluate the robustness of the structured reasoning framework against the inherent uncertainty in LLM output, we designed a set of experiments to examine whether variability during the reasoning phase of LLM affects the overall performance of CircuGraphRAG. Although the system leverages template matching and SPARQL query construction to enable structured question answering supported by a knowledge graph, the generation of these structures is still fundamentally dependent on the reasoning capability of the LLM. We hypothesize that if the LLM exhibits instability in template recognition or query path merging, it may lead to downstream retrieval failures or incorrect paths, ultimately degrading the answer accuracy.

We designed two sub-experiments, and to ensure experimental controllability, we conducted the evaluation on a representative single-hop question answering case. The first part evaluates the structural output consistency of the LLM in two key stages: \textit{template matching} and \textit{query merging}. Specifically, we compute three metrics over five repeated runs: (1) \textbf{ROUGE-L} for lexical similarity, (2) \textbf{BERTScore} for semantic similarity, and (3) \textbf{Exact Match Rate} to measure whether the generated structures remain identical across runs. The second part examines whether structural generation differences affect the final answer quality. We record the answer accuracy of CircuGraphRAG, Naive RAG, and Standalone LLM over five runs for each input and compare their accuracy levels.

\begin{figure}[ht]
    \centering
    \begin{subfigure}[t]{\textwidth}
        \centering
        \includegraphics[width=\textwidth]{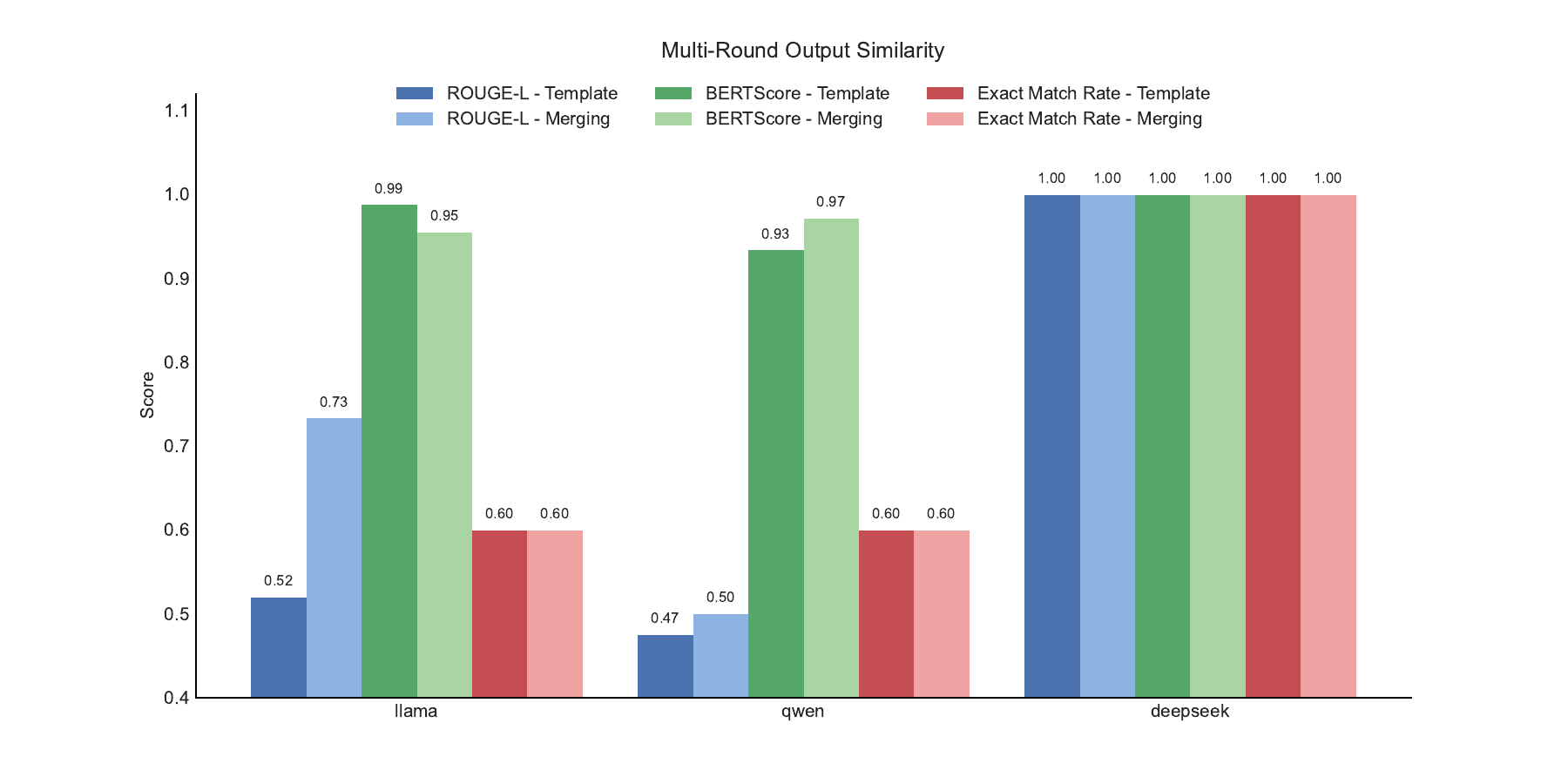}
        \caption{Multi-round output similarity across LLMs and stages.}
        \label{fig:multi-round-similarity}
    \end{subfigure}
    
    \vspace{1em} 
    
    \begin{subfigure}[t]{\textwidth}
        \centering
        \includegraphics[width=\textwidth]{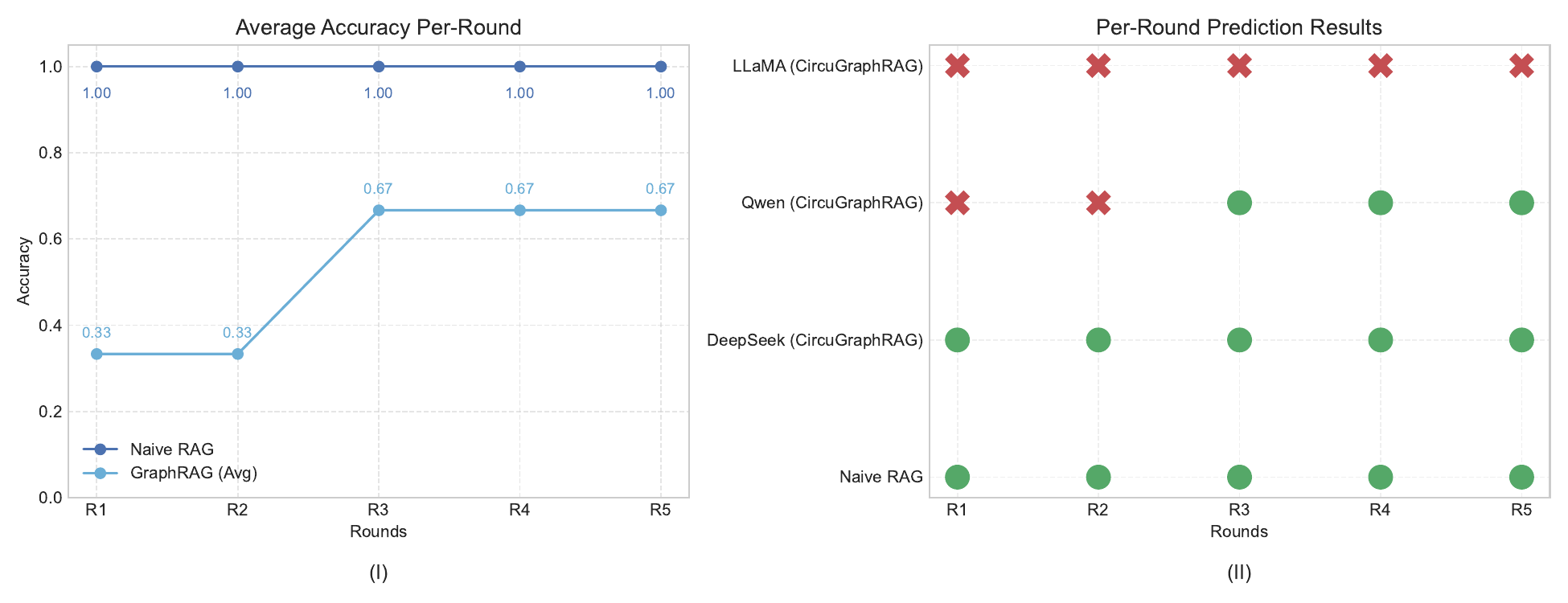}
        \caption{I. Average accuracy of CircuGraphRAG vs. Naive RAG over 5 rounds. II. Per-round correctness indicators for each model.}
        \label{fig:multi-round-accuracy-detail}
    \end{subfigure}
    
    \caption{Multi-round consistency and accuracy evaluation of CircuGraphRAG.}
    \label{fig:multi-round-combined}
\end{figure}


According to \autoref{fig:multi-round-combined}, the results confirm that LLMs exhibit significant instability in two critical stages of structured generation, template matching and query merging. Although DeepSeek maintains complete consistency across multiple reasoning rounds, both Llama and Qwen achieve only $60\%$ exact match rates in both stages, indicating substantial variability in symbolic output structures (as reflected in lower ROUGE-L scores). Although the semantic content remains largely consistent (as shown by high BERTScores), such structural fluctuations can still lead to erroneous retrieval paths, ultimately compromising the accuracy of the final answers.

In contrast, non-structured RAG methods such as Naive RAG and Standalone LLM demonstrate greater tolerance to reasoning variance, as they are not bound by rigid structural dependencies. This highlights a key trade-off: while structured QA frameworks enhance interpretability and control, they also amplify the adverse impact of LLM reasoning instability.

\subsection{Error Analysis and Qualitative Observations}

While CircuGraphRAG substantially reduces hallucinations and numeric errors, two primary limitations remain:
\begin{itemize}
    \item \textbf{Knowledge Gaps:} Some specialized or newly introduced waste streams may be missing in the knowledge graph. In such cases, CircuGraphRAG returns partial or incomplete results.
    \item \textbf{Vague User Prompts:} Overly broad queries can mismatch the available templates, producing off-target or partial retrieval. By contrast, standalone LLMs and naive RAG often produce confident but incorrect code assignments or invented numerical values.
\end{itemize}
These limitations underscore the importance of regularly updating the ontology and refining domain-specific query templates to ensure coverage of emerging regulations and precise user intent.

\subsection{Summary of Findings}

Our experiments demonstrate that aligning an LLM with a structured knowledge graph offers significant accuracy and efficiency gains for waste-to-resource management. CircuGraphRAG consistently delivers higher ROUGE-L scores in single and multihop tasks, reduces hallucinations and numeric inaccuracies, and in many queries lowers the overall token footprint by 15--16\% compared to naive retrieval enhancement. Its strong synergistic detection performance and precise code alignments highlight the potential of the system to navigate complex industrial ecology regulations and accelerate the real-world adoption of circular economy practices.

%% file: Sections/3-discussion.tex
\section{Discussion}
\label{sec:discussion}

Our findings illustrate that linking an LLM to a structured knowledge graph significantly improves accuracy, numeric consistency, and code alignment in waste-to-resource management. The examples presented in Appendix~\ref{sec:singlehop} and Appendix~\ref{sec:multihop} highlight how CircuGraphRAG outperforms both standalone LLMs and naive retrieval-augmented systems in representative queries, ranging from simple lookups from ``identifying the correct category for polyurethane waste'' to multi-hop reasoning such as ``matching NACE code 3821 to CPA code 382150''.

\subsection{Key Advances}

\subsubsection{Reduced Hallucinations in Single-Hop Queries}
We selected the classification of “waste polyurethane” as a representative single-hop query to evaluate the reasoning performance of CircuGraphRAG. While the ground truth specifies “construction and demolition wastes,” Standalone LLMs frequently produced overgeneralized or inconsistent categories, such as “municipal solid waste.” In contrast, CircuGraphRAG accurately identified the correct classification through its graph-based retrieval mechanism, which links “waste polyurethane” to its corresponding relationship (e.g., \texttt{hasCategory}) in the knowledge graph. In real-world applications, classification codes are often governed by regulatory standards for safe disposal or recycling, making the reduction of such hallucinations essential. Detailed results for this case are presented in Appendix~\ref{sec:singlehop}.

\subsubsection{Enhanced Multi-Hop Reasoning for the Identification of Synergies}
Multi-hop queries often involve combining multiple relationships, such as mapping a provider's NACE code (e.g., 3821) to the CPA code corresponding to its generated resources (e.g., 382150). CircuGraphRAG achieves exact matches in such scenarios, where both Standalone LLMs and naive retrieval methods typically fail, underscoring the system’s capacity to \emph{verify each step} against structured ontology links. This capability to handle multi-hop reasoning is essential for identifying opportunities for industrial symbiosis, such as matching metal slags from steelmaking with cement production, thereby supporting circular-economy initiatives. Illustrative multi-hop examples can be found in Appendix~\ref{sec:multihop}.

\subsubsection{Traceable Explanations and Regulatory Relevance}
Beyond numeric metrics, CircuGraphRAG ensures that each retrieval step is transparent, as illustrated in the ``Query Merging'' blocks in Appendix~\ref{sec:singlehop} and Appendix~\ref{sec:multihop}. Domain experts and regulators can audit exactly how the system arrived at a particular classification code or emission factor. Given the potential legal and environmental ramifications of an incorrect or fabricated classification (e.g., labeling a hazardous waste as non-hazardous), this built-in transparency meets a critical need in regulated industries.

\subsection{Practical Implications}

The improved accuracy and explainability of the system address key challenges in industrial ecology.

\subsubsection{Faster and More Reliable Synergy Discovery}
Appendix~\ref{sec:singlehop} and Appendix~\ref{sec:multihop} feature typical queries in which organizations seek to identify compatible receivers for by-products. CircuGraphRAG reduces the manual overhead of cross-referencing waste codes and emission thresholds, thus surfacing reuse pathways that a purely text-based approach might overlook.

\subsubsection{Minimizing Incorrect Classifications}
Even minor classification errors (e.g., mislabeling polyurethane as municipal waste) can lead to non-compliance or suboptimal resource routing. The examples show that CircuGraphRAG’s graph-anchored retrieval routinely supplies the correct code alignment, enabling more rigorous decision-making for facility-level or regional-scale waste management.

By systematically linking classification codes, GWP100, and facility attributes, CircuGraphRAG accelerates large-scale circular initiatives. Its subgraph retrieval also provides an auditable trail that reassures regulators and other stakeholders, who can confirm that each proposal meets legal or emissions requirements before implementation.

%% file: Sections/4-methods.tex
\section{Methods}
\label{sec:method}
We propose CircuGraphRAG, a retrieval-augmented LLM framework that integrates multiple industrial classification systems, environmental impact factors, and textual embeddings into a unified knowledge graph. By automating single-hop and multi-hop queries, CircuGraphRAG identifies feasible ``waste-to-resource'' routes across industries, highlighting those with low GWP100 emissions. Our approach tackles two core challenges of sustainable industrial ecology: 
\begin{itemize}
    \item the diversity and complexity of classification schemes for waste and industrial processes, and 
    \item the need to ground LLM responses in reliable, structured data to avoid hallucinations.
\end{itemize}
The following sections describe each methodological step, from dataset construction to the final synergy recommendations ranked by GWP100.

\subsection{Dataset Overview}
\label{subsec:dataset-kg}

Our experiments make use of 3{,}896 entries from the ``Waste Treatment and Recycling'' sector of the Ecoinvent dataset~\cite{FRISCHKNECHT20051337}, each mapped to International Standard Industrial Classification of All Economic Activities (ISIC), Central Product Classification (CPC), Nomenclature statistique des activités économiques dans la Communauté européenne (NACE), Wirtschaftszweige (WZ),  Classification of Products by Activity (CPA), European Waste Catalogue (EWC), Harmonized System (HS), and Singapore Standard Industrial Classification (SSIC) codes, as well as the respective impact factor GWP100. We merge domain-specific information such as facility locations, emissions data, provider and receiver identities, and regulatory constraints into a unified knowledge graph. This preserves the original code mappings and introduces semantic links across datasets. In practice, we harmonize codes from multiple classification systems to ensure cross-regional compatibility. The resulting knowledge graph includes the following:
\begin{itemize}
    \item \textbf{117,380 nodes}, each representing a discrete resource entity (e.g., waste streams, by-products, industrial facilities, emission flows and regulatory guidelines).
    \item \textbf{753,145 edges}, capturing relationships such as \texttt{hasGwp100} (mapping a resource to its \emph{GWP100}), \texttt{hasProvider}, \texttt{hasReceiver} (linking facilities that generate or accept a given waste stream), and \texttt{hasResource} (connecting a by-product to relevant industrial processes).
    \item An \textbf{integrated ontology} that consolidates EWC, NACE, ISIC, and SSIC coding, normalizing terminology that would otherwise vary across geographies or regulatory bodies.
\end{itemize}

\subsection{Dataset Construction}
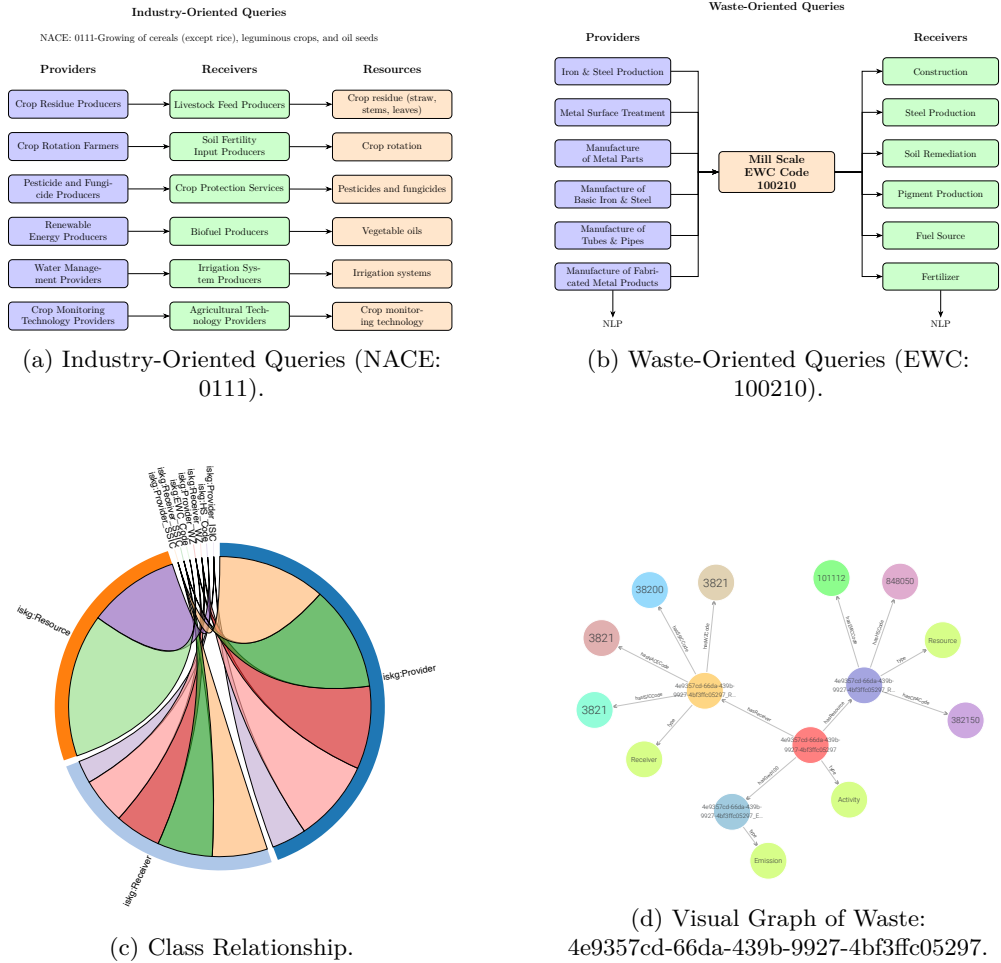
\begin{figure}[!ht]
\centering

\begin{subfigure}[b]{0.45\textwidth}
    \centering
    \resizebox{\linewidth}{!}{%
    \begin{tikzpicture}[
        node distance=0.5cm and 1.5cm,
        box/.style={rectangle, draw, fill=#1!20, 
                    text width=4cm, align=center, 
                    rounded corners, minimum height=1cm},
        arrow/.style={-Stealth, thick},
        title/.style={font=\bfseries\large}
    ]
    \node[title] (title) at (0,0) {Industry-Oriented Queries};
    \node[align=center, text width=12cm, below=0.3cm of title] (subtitle) 
        {NACE: 0111-Growing of cereals (except rice), leguminous crops, and oil seeds};
    \node[title, below=0.5cm of subtitle, xshift=-5cm] (providers) {Providers};
    \node[title, right=of providers, xshift=2cm] (receivers) {Receivers};
    \node[title, right=of receivers, xshift=2cm] (resources) {Resources};
    \node[box=blue, below=0.5cm of providers] (p1) {Crop Residue Producers};
    \node[box=blue, below=0.5cm of p1] (p2) {Crop Rotation Farmers};
    \node[box=blue, below=0.5cm of p2] (p3) {Pesticide and Fungicide Producers};
    \node[box=blue, below=0.5cm of p3] (p4) {Renewable Energy Producers};
    \node[box=blue, below=0.5cm of p4] (p5) {Water Management Providers};
    \node[box=blue, below=0.5cm of p5] (p6) {Crop Monitoring Technology Providers};
    \node[box=green, below=0.5cm of receivers] (r1) {Livestock Feed Producers};
    \node[box=green, below=0.5cm of r1] (r2) {Soil Fertility Input Producers};
    \node[box=green, below=0.5cm of r2] (r3) {Crop Protection Services};
    \node[box=green, below=0.5cm of r3] (r4) {Biofuel Producers};
    \node[box=green, below=0.5cm of r4] (r5) {Irrigation System Producers};
    \node[box=green, below=0.5cm of r5] (r6) {Agricultural Technology Providers};
    \node[box=orange, below=0.5cm of resources] (res1) {Crop residue (straw, stems, leaves)};
    \node[box=orange, below=0.5cm of res1] (res2) {Crop rotation};
    \node[box=orange, below=0.5cm of res2] (res3) {Pesticides and fungicides};
    \node[box=orange, below=0.5cm of res3] (res4) {Vegetable oils};
    \node[box=orange, below=0.5cm of res4] (res5) {Irrigation systems};
    \node[box=orange, below=0.5cm of res5] (res6) {Crop monitoring technology};

    \foreach \i in {1,...,6}
    {
        \draw[arrow] (p\i) -- (r\i);
        \draw[arrow] (r\i) -- (res\i);
    }
    \end{tikzpicture}
    } 
    \caption{Industry-Oriented Queries (NACE: 0111).}
    \label{fig:ioq}
\end{subfigure}
\hfill
\begin{subfigure}[b]{0.45\textwidth}
    \centering
    \resizebox{\linewidth}{!}{%
    \begin{tikzpicture}[
        node distance=0.5cm and 1.5cm,
        box/.style={rectangle, draw, fill=#1!20, 
                    text width=4cm, align=center, 
                    rounded corners, minimum height=1cm},
        arrow/.style={-Stealth, thick},
        title/.style={font=\bfseries\large}
    ]
    \node[title] (title) at (0,0) {Waste-Oriented Queries};
    \node[box=orange, title, below=5cm of title] (central) 
        {Mill Scale\\EWC Code\\100210};
    \node[title, below=0.5cm of title, xshift=-6cm] (providers) {Providers};
    \node[title, below=0.5cm of title, xshift=6cm] (receivers) {Receivers};
    \node[box=blue, below=0.5cm of providers] (p1) {Iron \& Steel Production};
    \node[box=blue, below=0.5cm of p1] (p2) {Metal Surface Treatment};
    \node[box=blue, below=0.5cm of p2] (p3) {Manufacture of Metal Parts};
    \node[box=blue, below=0.5cm of p3] (p4) {Manufacture of Basic Iron \& Steel};
    \node[box=blue, below=0.5cm of p4] (p5) {Manufacture of Tubes \& Pipes};
    \node[box=blue, below=0.5cm of p5] (p6) {Manufacture of Fabricated Metal Products};
    \node[box=green, below=0.5cm of receivers] (r1) {Construction};
    \node[box=green, below=0.5cm of r1] (r2) {Steel Production};
    \node[box=green, below=0.5cm of r2] (r3) {Soil Remediation};
    \node[box=green, below=0.5cm of r3] (r4) {Pigment Production};
    \node[box=green, below=0.5cm of r4] (r5) {Fuel Source};
    \node[box=green, below=0.5cm of r5] (r6) {Fertilizer};

    \foreach \i in {1,...,6}
    {
        \draw[arrow] (p\i.east) -- ++(1,0) |- (central.west);
    }
    \foreach \i in {1,...,6}
    {
        \draw[arrow] (central.east) -- ++(1,0) |- (r\i.west);
    }
    
    \node[align=center, below=1cm of p6] (nlp1) {NLP};
    \node[align=center, below=1cm of r6] (nlp2) {NLP};
    \draw[arrow] (p6.south) -- (nlp1.north);
    \draw[arrow] (r6.south) -- (nlp2.north);
    \end{tikzpicture}
    } 
    \caption{Waste-Oriented Queries (EWC: 100210).}
    \label{fig:woq}
\end{subfigure}

\vskip\baselineskip 

\begin{subfigure}[b]{0.45\textwidth}
    \centering
    \resizebox{\linewidth}{!}{%
        \includegraphics{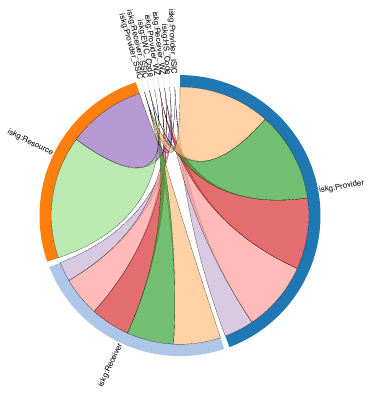}
    }
    \caption{Class Relationship.}
    \label{fig:class_relationships}
\end{subfigure}
\hfill
\begin{subfigure}[b]{0.45\textwidth}
    \centering
    \resizebox{\linewidth}{!}{%
        \includegraphics{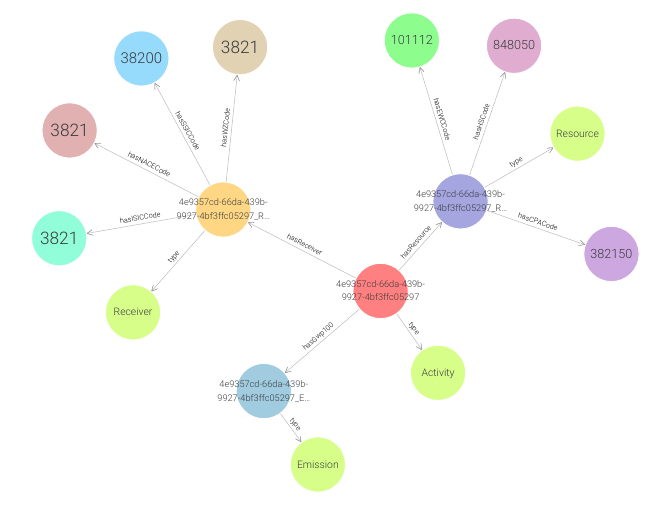}
    }
    \caption{Visual Graph of Waste: 4e9357cd-66da-439b-9927-4bf3ffc05297.}
    \label{fig:visual-graph}
\end{subfigure}

\caption{An overview of the structure of our knowledge graph database.}
\label{fig:combined}
\end{figure}

We draw on the Ecoinvent database~\cite{FRISCHKNECHT20051337}, a comprehensive repository of industrial processes, resource flows, and emissions data as shown in \autoref{fig:combined}. Each record is enriched with standard classification codes, such as ISIC, NACE, SSIC, WZ, EWC, HS, and CPA, thus enabling both \emph{industry-oriented} and \emph{waste-oriented} analyses:
\begin{itemize}
    \item \textbf{Industry-oriented queries:} Identify synergies between industries based on shared or compatible classification codes, as shown in \autoref{fig:ioq}.
    \item \textbf{Waste-oriented queries:} Pinpoint potential receivers for a given waste stream aligned with regulatory codes and environmental metrics, as shown in \autoref{fig:woq}.
\end{itemize}
Where direct crosswalks are incomplete, we apply sentence-embedding techniques (\texttt{multi-qa-mpnet-base-cos-v1}) to infer plausible mappings from textual descriptions. Newly inferred codes are flagged for domain experts to validate. This harmonized dataset provides a robust basis for constructing our knowledge graph, ensuring consistent references for industrial activities and waste categories across multiple sectors and jurisdictions.

\subsection{Ground Truth Dataset Construction}
\label{subsec:ground-truth}

To rigorously measure the accuracy of the response, we developed a set of canonical queries and the corresponding \emph{reference ground truth}. \autoref{tab:groundtruth-examples} illustrates the variety of single-hop and multi-hop queries used to evaluate the systems. Each query--answer pair underwent a multi-hop curation process to ensure industrial relevance and technical correctness:
\begin{enumerate}
    \item \textbf{Query Selection:} We identified representative questions reflecting real-world needs in waste-to-resource management, including code alignment (EWC, HS, NACE, CPA) and key attributes (GWP100). These queries span single-hop lookups (e.g., retrieving a code-specific entity) and multi-hop reasoning (e.g., applying regulatory constraints across multiple entities).
    \item \textbf{Domain Annotation:} We cross-referenced official coding guidelines and our expanded knowledge graph to establish gold-standard answers. Verification was centered on correct waste code matching, valid provider--receiver relationships, and consistent metadata.
    \item \textbf{Cross-Checking:} Each annotated answer was tested against our knowledge graph to confirm that the declared relationships were unambiguously retrievable. Any inconsistencies resulted in further refinements to the knowledge graph or the ground truth.
\end{enumerate}

Subsequently, we evaluated the performance of our proposed CircuGraphRAG approach on two primary tasks: answering \emph{single-hop} and \emph{multi-hop} queries. These tasks range from straightforward fact retrieval, such as obtaining a GWP100 value, to multi-hop reasoning that involves identifying by-products and matching them with suitable receivers under specific regulatory constraints. We compare CircuGraphRAG with two baselines:

\begin{enumerate}
    \item \textbf{Standalone LLM:} An LLM in isolation (Llama, Qwen, DeepSeek) without explicit retrieval.
    \item \textbf{Naive RAG:} A RAG approach that lacks fine-grained graph constraints.
\end{enumerate}

\subsection{Knowledge Graph Construction}
\begin{enumerate}
    \item \textbf{Providers:} Entities or processes that generate secondary materials or by-products.
    \item \textbf{Receivers:} Facilities that reuse or recycle these materials.
    \item \textbf{Resources:} The actual material flows, annotated with attributes such as GWP100 or hazardous status.
\end{enumerate}
Properties (e.g., \texttt{iskg:hasEWCCode}, \texttt{iskg:hasGwp100}) link these classes as shown in \autoref{fig:class_relationships}, enabling SPARQL queries for fine-grained investigations of waste-to-resource pathways. Using canonical URIs for classification codes, we maintain semantic clarity even when bridging multiple standards (ISIC, NACE, EWC, etc.) as shown in \autoref{fig:visual-graph}.

\subsection{The CircuGraphRAG Framework}
\label{subsec:circu-graph-rag}

\begin{figure}[!ht]
    \centering
    \includegraphics[width=\linewidth]{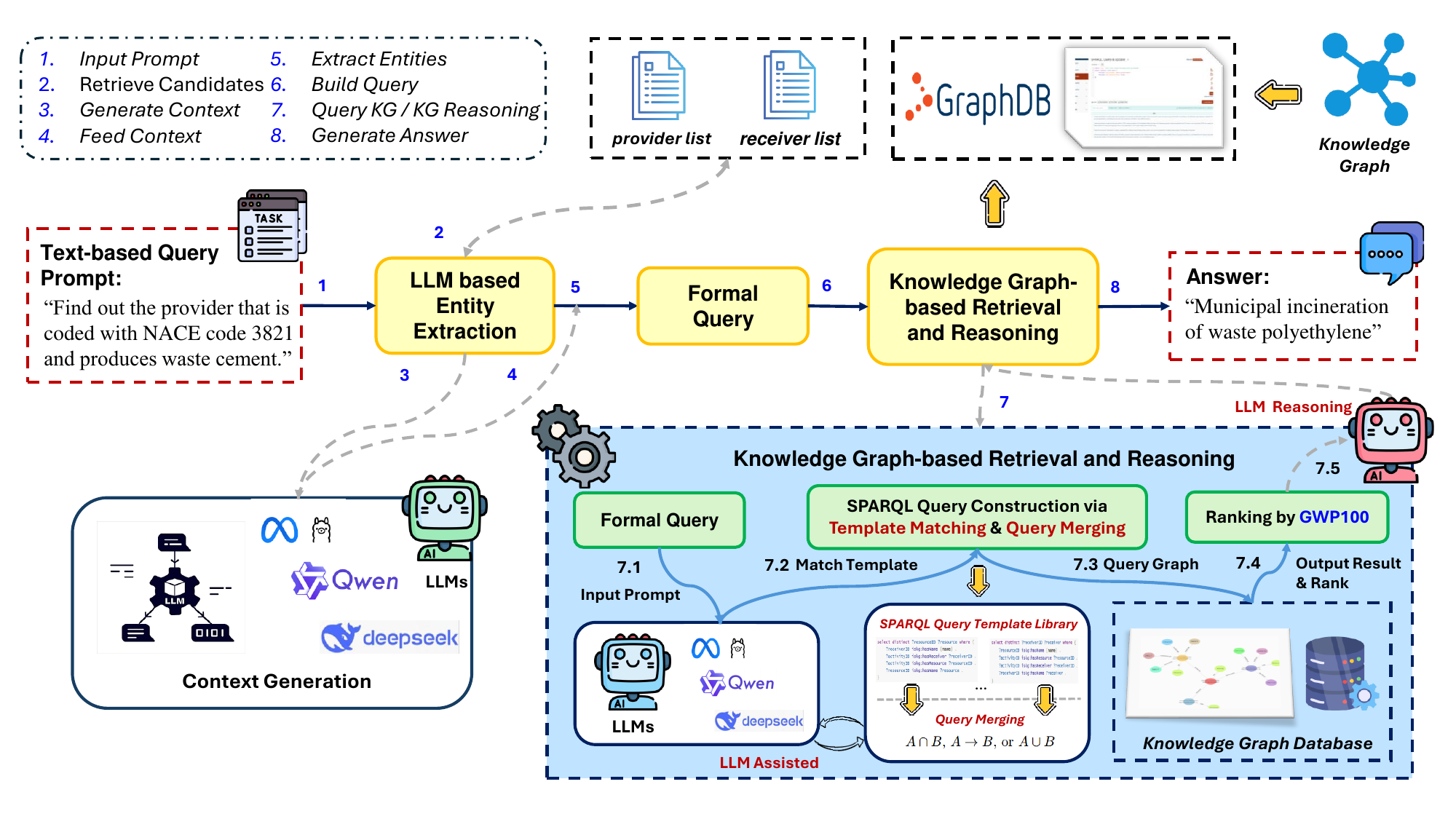}
    \caption{The CircuGraphRAG Pipeline.}
    \label{fig:graphrag_pipeline}
\end{figure}

CircuGraphRAG unifies knowledge graph queries with large language models in a retrieval-augmented generation pipeline (see \autoref{fig:graphrag_pipeline}). Its main components include :

\subsubsection{Text-Based Query}

Users submit queries in free text form, often specifying a particular industrial code or waste type (e.g. ``Find the provider with NACE code 3821 producing waste cement"). An LLM then parses this prompt to extract key concepts (e.g., “NACE 3821,” “waste cement”).

\subsubsection{Entity Classification \& Vector Indexing}
Each user query is parsed to determine whether it targets a \emph{provider} or a \emph{receiver}. Potential entities are encoded into 384-dimensional embeddings using \texttt{all-MiniLM-L6-v2} and stored in two Facebook AI Similarity Search (FAISS) indexes (one for providers, one for receivers). For each query embedding $\mathbf{q} \in \mathbb{R}^d$, we retrieve the top-$k$ most similar entity vectors $\{\mathbf{e}_1, \dots, \mathbf{e}_k\}$ by maximizing the inner product
\[
\text{score}(\mathbf{q}, \mathbf{e}_i) = \langle \mathbf{q}, \mathbf{e}_i \rangle = \sum_{j=1}^{d} q_j \cdot e_{i,j}.
\]
This is implemented efficiently using \texttt{faiss.IndexFlatIP}, which ranks all entity vectors in descending order of similarity. Top-$k$ candidates are retrieved for downstream reasoning and reranking.

\subsubsection{Knowledge Graph-based Retrieval and Reasoning}

Based on Algorithm~\ref{alg:spaqrl-qa}, the knowledge graph-based retrieval process uses an LLM-driven template matcher to interpret user queries and produce appropriate SPARQL statements. We maintain a TemplateLibrary of 18 distinct SPARQL templates, each designed to address common query intentions (for example, “Find all receivers for a specific waste stream,” “Locate facilities by classification code,” etc.). In Step~1, the LLM identifies relevant entities (for example, a facility code or the name of a resource) and selects all templates that match the semantic intent of the user’s question. These matching templates, often more than one, have placeholders (such as \texttt{?providerCode} or \texttt{?resourceName}) that are filled with information extracted from the query.

In Step 2, if multiple templates are selected, the LLM determines how to merge them according to the query requirements. Depending on whether the user needs intersecting results, chained filtering, or broader coverage, the templates are combined using operations, including intersection, chaining, or union. This automated merging ensures that all relevant aspects of the user’s request are handled in a single FinalQuery.

Subsequently, in Step~3, we execute the FinalQuery against the knowledge graph and retrieve a structured result set that contains facts such as the matching of facility names, classification codes, and values GWP100. We then rank the results by ascending GWP100, ensuring that the most environmentally favorable reuse pathways appear at the top. Typically, we return the top-5 results by default, but the user can specify any subset size or additional filters as needed.

\begin{algorithm}[!ht]
\caption{Knowledge Graph-based Retrieval and Reasoning Algorithm}
\label{alg:spaqrl-qa}
\begin{algorithmic}[1]

\Require A natural language question $Q$
\Ensure A concise answer to $Q$

\State \textbf{Step 1: Template Matching (LLM-Reasoning)}
\State Use an LLM to parse $Q$ and identify relevant entities (e.g., codes, resource or provider names), along with the intended search objective.
\State \textbf{MatchedTemplates} $\gets \{\}$ 
\ForAll{\texttt{SPARQL template } $T$ in \textbf{TemplateLibrary}}
  \If{\texttt{LLM semantic matching} confirms $T$ fits $Q$'s intent}
    \State \textbf{MatchedTemplates} $\leftarrow \textbf{MatchedTemplates} \cup 
           \{(\textit{templateID}=T,\ \textit{inputFields from }Q,\ \textit{expectedOutput})\}$
  \EndIf
\EndFor

\State

\State \textbf{Step 2: Query Merging (LLM-Reasoning)}
\State If multiple templates are selected, use an LLM to determine the optimal merging methods 
       (e.g., intersection, chaining, union) based on $Q$'s requirements.
\State Combine all templates in \textbf{MatchedTemplates} into a single SPARQL query 
      \textbf{FinalQuery} (e.g., $A \cap B$, $A \to B$, or $A \cup B$).

\State

\State \textbf{Step 3: Execute and Retrieve Answer}
\State Execute \textbf{FinalQuery} against the knowledge graph to obtain the result set $\mathbf{R}$.
\State Parse $\mathbf{R}$ to extract the entities or values requested by $Q$.
\State \textbf{return} a concise answer derived from $\mathbf{R}$.

\end{algorithmic}
\end{algorithm}

\subsubsection{LLM Response Composition}

Finally, a concise LLM generated response is displayed to the user. Because LLM output is tightly grounded in knowledge graph data (rather than raw text fragments), hallucination is greatly reduced. Each recommendation can be traced back to its matching entities, classification codes, and GWP100 values on the RDF graph, promoting transparency and reproducibility.

If the knowledge graph contains no relevant entries or mappings for the user’s query, the system returns a brief message indicating that no matches were found in the database. However, the pipeline then relies on the broader knowledge of the LLM to provide any contextual information available, acknowledging that these details are not drawn from the verified graph. This fallback approach ensures that the user receives at least a partial or approximate answer, while clearly distinguishing graph‐based facts from the LLM’s more general training data. Users can refine the query or add additional details to improve the precision of future lookups and guide the continuous expansion of the knowledge graph coverage.



\subsection{Experimental Setup}

\subsubsection{Hardware and Environment}
All experiments are conducted on a Linux server equipped with an NVIDIA 4090\,GPU, dual Intel Xeon processors, and 256\,GB of RAM. We host the knowledge graph using GraphDB, and handle SPARQL queries through \texttt{SPARQLWrapper}. Vector embeddings are managed by FAISS, while LLM inference is provided via the Groq API. To enable flexible deployments across different environments, we use Python's \texttt{dotenv} to store endpoint URLs and credentials.

\subsubsection{Models Evaluated}
We benchmark the following models:

\begin{itemize}
    \item \textbf{Llama3 (70B)}~\cite{meta2024llama3}: A 70B-parameter transformer model with a 4{,}096-token context window.
    \item \textbf{Qwen-qwq (32B)}~\cite{qwen2024qwq32b}: A 32B-parameter variant tailored for domain-oriented tasks.
    \item \textbf{DeepSeek-R1-Distill-Llama (70B)}~\cite{deepseek2025r1}: A 70B-parameter Llama derivative fine-tuned on specialized corpora.
\end{itemize}

All three models share identical inference hyperparameters (\texttt{temperature} = 0.7, \texttt{top-p} = 0.9, \texttt{max\_tokens} = 4096) unless otherwise specified.

\subsubsection{Baselines and Proposed Method}
We compare our proposed CircuGraphRAG system to two baselines:
\begin{enumerate}
    \item \textbf{Standalone LLM}: The user query is directly passed to the model without any external retrieval step.
    \item \textbf{Naive RAG}: Unstructured text segments are retrieved, rather than relying on SPARQL queries and a structured knowledge graph.
\end{enumerate}

In contrast, CircuGraphRAG combines structured SPARQL-based retrieval with a knowledge graph to improve both factual precision and sustainability-focused relevance.

\subsection{Evaluation Metrics}

\subsubsection{Query Types}

\paragraph{1. Single-Hop Queries}
Examines whether the system can accurately retrieve a single fact from the graph.

\paragraph{2. Multi-Hop Queries}
Determines the capacity of the system to chain multiple codes, constraints, and industrial relationships for valid reuse pathways.

\subsubsection{Accuracy}
We report two types of accuracy to evaluate the consistency of multi-round outputs and the correctness of responses under different template matching settings.

\paragraph{1. Round-level Precision}
Measures the average accuracy of the model output over a sequence of evaluation rounds (Figure~\ref{fig:multi-round-accuracy-detail}). It is defined as
\[
\text{Accuracy}_{\text{round}} 
= \frac{1}{T} \sum_{t=1}^{T} \mathbb{1}[y_t^{\text{output}} = y_t^{\text{true}}],
\tag{1}
\]
where \( T \) is the total number of rounds.

\paragraph{2. Template Matching Accuracy}
Assesses whether the system generates correct outputs under different template matching settings (Table~\ref{tab:template-ablation}). 

\subsubsection{ROUGE-L Precision, Recall, and F1}
We use ROUGE-L, a recall-oriented metric based on the Longest Common Subsequence (LCS), to evaluate the lexical overlap between the generated and reference answers. 

\paragraph{1. ROUGE-L Precision} 
Measures how much of the generated answer matches the reference, relative to the generated length:
\[
\text{Precision}_{\text{ROUGE-L}} 
= \frac{LCS(C, R)}{|C|}.
\tag{2}
\]

\paragraph{2. ROUGE-L Recall} 
Measures how much of the reference answer is captured in the generated answer:
\[
\text{Recall}_{\text{ROUGE-L}} 
= \frac{LCS(C, R)}{|R|}.
\tag{3}
\]

\paragraph{3. ROUGE-L F1 Score} 
Is the harmonic mean of ROUGE-L precision and recall:
\[
\text{F1}_{\text{ROUGE-L}} 
= \frac{(1 + \beta^2) \cdot \text{Precision}_{\text{ROUGE-L}} \cdot \text{Recall}_{\text{ROUGE-L}}}
{\text{Precision}_{\text{ROUGE-L}} + \beta^2 \cdot \text{Recall}_{\text{ROUGE-L}}}.
\tag{4}
\]
where \( \beta = 1 \) by default. Here, \( LCS(C, R) \) denotes the length of the longest common subsequence between the candidate answer \( C \) and the reference answer \( R \).

\subsubsection{BERTScore}
BERTScore evaluates the semantic similarity between a generated answer 
\( C = \{c_1, \dots, c_m\} \) 
and a reference answer 
\( R = \{r_1, \dots, r_n\} \)
by computing pairwise cosine similarities between their contextual embeddings. The final score is
\[
\text{BERTScore}(C, R) 
= \frac{1}{2} 
\Biggl( \frac{1}{|C|} \sum_{c_i \in C} \max_{r_j \in R} \cos(\vec{c}_i, \vec{r}_j)
+ \frac{1}{|R|} \sum_{r_j \in R} \max_{c_i \in C} \cos(\vec{r}_j, \vec{c}_i)
\Biggr),
\tag{5}
\]
where \( \cos(\vec{c}_i, \vec{r}_j) \) denotes the cosine similarity between the contextual embeddings of the tokens \( c_i \) and \( r_j \).

\subsubsection{Exact Match Rate}
EM measures the strictest form of matching, evaluating whether the predicted answer exactly matches the reference answer after normalization (lowercasing, removing articles and punctuation, etc.).
\[
\text{EM} 
= \frac{1}{N} 
\sum_{i=1}^{N} \mathbb{1}[\hat{y}_i = y_i],
\tag{6}
\]
where \( \hat{y}_i \) is the predicted answer, \( y_i \) is the reference answer, \( \mathbb{1}[\cdot] \) is the indicator function and \( N \) is the total number of evaluation instances.

\noindent
By presenting the query process in a knowledge graph and ranking synergies by GWP100, CircuGraphRAG improves both the factual precision and environmental relevance of “waste-to-resource” recommendations, thus advancing data-driven sustainability in industrial ecosystems.

%% file: Sections/5-conclusion.tex
\section{Conclusion}
\label{sec:conclusion}

We introduced CircuGraphRAG, a knowledge graph based RAG framework that unites knowledge graph reasoning with LLMs to address waste-to-resource challenges in the circular economy. By mapping diverse industrial codes (EWC, NACE, ISIC, SSIC) and embedding regulatory metadata into a unified graph, CircuGraphRAG consistently surpassed baseline methods on single- and multi-hop queries, reducing hallucinations
and boosting synergy discovery among potential waste producers and reusers. 

Our experiments showed clear gains in factual accuracy, improved multi-hop reasoning, and lowered token usage when retrieving only relevant subgraphs. However, the effectiveness of CircuGraphRAG relies on continuously updated ontologies and access to real-time data, both of which remain open challenges. Future work will explore expanding domain coverage, incorporating dynamic sensor data, and collaborating with policymakers to encourage standardized data reporting. By systematically connecting industrial by-products to potential receivers, CircuGraphRAG marks a promising step toward more sustainable, circular production systems. 

\section*{Declarations}

\begin{itemize}
\item\textbf{Conflict of interest/Competing interests:} 
The authors declare that they have no conflict of interest. 

\item\textbf{Ethics approval and consent to participate:} 
Not applicable. 

\item\textbf{Consent for publication:} 
Not applicable. 

\item\textbf{Data availability:}
Part of the data used in this study is obtained from the Ecoinvent dataset~\cite{FRISCHKNECHT20051337}. In compliance with data licensing terms, no actual numerical values from Ecoinvent are disclosed in this paper. The remaining datasets will be publicly released upon publication.

\item\textbf{Materials availability:} 
Not applicable. 

\item\textbf{Code availability:} 
The relevant code for this study will be made publicly available after publication

\end{itemize}

%% file: Sections/6-appendix.tex
\newpage
\section{Appendix}
\label{sec:appendix}

\subsection{Query--Answer Pairs from the Ground Truth.}
\label{sec:groundtruth-examples}
\begin{table}[ht!]
\centering
\caption{Query--Answer Pairs from the Ground Truth.}
\label{tab:groundtruth-examples}
\begin{tabular}{p{0.5cm} p{1.5cm} p{6cm} p{4cm}}
\toprule
\textbf{Case} & \textbf{Type} & \textbf{Query} & \textbf{Answer} \\
\midrule
1 & Single-Hop 
  & \emph{Q: Find out the resource that is coded with EWC code 080121 and HS code 810330.}  
  & A: Waste paint \\ \hline
2 & Single-Hop 
  & \emph{Q: Find out the provider that is coded with NACE code 3821 and produces waste cement.}  
  & A: Municipal incineration of waste polyethylene \\ \hline
3 & Single-Hop 
  & \emph{Q: Find out the least value of GWP100 among the receivers coded with NACE code 3822.}
  & A: 0.008826959 \\ \hline
4 & Multi-Hop 
  & \emph{Q: The resources generated by the providers under the NACE code 3821 are all categorized under what CPA code and category?}
  & A: 382150; ``Pellets of municipal waste'' \\ \hline
5 & Multi-Hop 
  & \emph{Q: Among the receivers of resources coded with CPA code 382150, which can be matched again with the receiver ``market for waste polystyrene'' as a provider?}
  & A: ``treatment of waste polystyrene terephthalate, municipal incineration'' \\ \hline
6 & Multi-Hop 
  & \emph{Q: Among the providers of resources coded with EWC code 070213, which can also be a receiver of the ``aluminium scrap''?}
  & A: ``treatment of wastewater from PV cell production, wastewater treatment'' \\
\bottomrule
\end{tabular}
\end{table}

\newpage
\subsection{Single-hop QA performance for different LLMs and retrieval methods across 3 test cases.}
\label{sec:single-hop-metrics}
\begin{table}[!ht]
\caption{Single-hop QA performance for different LLMs and retrieval methods across 3 test cases.}
\label{single-hop-metrics}
\begin{tabular}{llllll}
\hline
\multirow{2}{*}{Case} & \multirow{2}{*}{Method}     & \multirow{2}{*}{LLM} & \multicolumn{3}{c}{ROUGE-L} \\ \cmidrule{4-6} 
                                 &                             &                      & Precision & Recall & F1     \\ \hline
\multirow{9}{*}{1}          & \multirow{3}{*}{Standalone LLM} & llama           &  0.0161      & 0.5 & 0.0313 \\ \cmidrule{3-6}
                                 &                             & qwen           &  0.0045      & 0.5 & 0.0090 \\ \cmidrule{3-6}
                                 &                             & deepseek             & 0.0086         & 0.5      & 0.0169      \\ \cmidrule{2-6}
                                 & \multirow{3}{*}{Naive RAG}  & llama          &  0      & 0 & 0 \\ \cmidrule{3-6}
                                 &                             & qwen           &  0      &0  & 0 \\ \cmidrule{3-6}
                                 &                             & deepseek             & 0         & 0      & 0      \\ \cmidrule{2-6}
                                 & \multirow{3}{*}{CircuGraphRAG}  & llama          & 0       & 0 & 0  \\ \cmidrule{3-6}
                                 &                             & qwen           & 1       & 1 & 1 \\ \cmidrule{3-6}
                                 &                             & deepseek             & 1         & 1      & 1      \\ \hline
\multirow{9}{*}{2}          & \multirow{3}{*}{Standalone LLM} & llama          &  0.0165       & 0.4 & 0.0317 \\ \cmidrule{3-6}
                                 &                             & qwen           &  0.0064      & 0.4 & 0.0127 \\ \cmidrule{3-6}
                                 &                             & deepseek             &  0.0213      & 0.4 & 0.0404 \\ \cmidrule{2-6}
                                 & \multirow{3}{*}{Naive RAG}                  & llama          &  0.0099      &  0.4 & 0.0192 \\ \cmidrule{3-6}
                                 &                             & qwen           &  0.0150      & 0.4 & 0.0290 \\ \cmidrule{3-6}
                                 &                             & deepseek             & 0.0230         & 0.4      & 0.0435      \\ \cmidrule{2-6}
                                 & \multirow{3}{*}{CircuGraphRAG}  & llama          & 0.4286       & 0.6 & 0.5 \\ \cmidrule{3-6}
                                 &                             & qwen           & 0.4286       & 0.6 & 0.5 \\ \cmidrule{3-6}
                                 &                             & deepseek             & 0.0909         & 0.2      & 0.125      \\ \hline
\multirow{5}{*}{3}          & \multirow{3}{*}{Standalone LLM} & llama          & 0         & 0      & 0      \\ \cmidrule{3-6}
                                 &                             & qwen           & 0.0025         & 0.5      & 0.0050      \\ \cmidrule{3-6}
                                 &                             & deepseek             & 0         & 0      & 0      \\ \cmidrule{2-6}
                                 & Naive RAG                   & llama          & 0         & 0      & 0      \\ \cmidrule{3-6}
                                &                             & qwen           &  0      &  0 & 0 \\ \cmidrule{3-6}
                                 &                             & deepseek             & 0         & 0      & 0      \\ \cmidrule{2-6}
                                 & CircuGraphRAG                    & llama          & 1       & 1 & 1 \\ \cmidrule{3-6}
                                &                             & qwen           & 1       & 1 & 1 \\ \cmidrule{3-6}
                                 &                             & deepseek             & 0         & 0      & 0      \\\hline
\end{tabular}
\end{table}

\newpage
\subsection{Multi-hop QA performance for different LLMs and retrieval methods across 3 test cases.}
\label{sec:multi-hop-metrics}
\begin{table}[!ht]
\caption{Multi-hop QA performance for different LLMs and retrieval methods across 3 test cases.}
\label{multi-hop-metrics}
\begin{tabular}{llllll}
\hline
\multirow{2}{*}{Case} & \multirow{2}{*}{Method}     & \multirow{2}{*}{LLM} & \multicolumn{3}{c}{ROUGE-L} \\ \cmidrule{4-6} 
                                 &                             &                      & Precision & Recall & F1     \\ \hline
\multirow{5}{*}{4}          & \multirow{3}{*}{Standalone LLM} & llama          &  0   &  0 & 0 \\ \cmidrule{3-6}
                                 &                             & qwen           &  0   &  0 & 0 \\ \cmidrule{3-6}
                                 &                             & deepseek             & 0        & 0     & 0      \\ \cmidrule{2-6}
                                 & \multirow{3}{*}{Naive RAG}                  & llama          &  0   &  0 & 0 \\ \cmidrule{3-6}
                                       &                             & qwen           &   0 &  0  & 0 \\ \cmidrule{3-6}
                                 &                             & deepseek             & 0         & 0      & 0      \\ \cmidrule{2-6}
                                 & \multirow{3}{*}{CircuGraphRAG}    & llama          &  0    & 0 & 0 \\ \cmidrule{3-6}
                                       &                             & qwen           &  1 &  1  & 1 \\ \cmidrule{3-6}
                                 &                             & deepseek             & 0         & 0      & 0      \\ \hline
\multirow{5}{*}{5}          & \multirow{3}{*}{Standalone LLM} & llama          &  0.0273     & 0.4286 & 0.0513 \\ \cmidrule{3-6}
                                 &                             & qwen           &  0.0385    & 0.2857 & 0.0678 \\ \cmidrule{3-6}
                                 &                             & deepseek             &  0.0303   & 0.4286  & 0.0566 \\ \cmidrule{2-6}
                                 & \multirow{3}{*}{Naive RAG}                   & llama          & 0.0160    &  0.2857 & 0.0303 \\ \cmidrule{3-6}
                                 &                             & qwen           &  0.0207    & 0.4286 & 0.0395 \\ \cmidrule{3-6}
                                 &                             & deepseek             &   0.0417  & 0.4286 & 0.0759 \\ \cmidrule{2-6}
                                 & \multirow{3}{*}{CircuGraphRAG}                    & llama          & 1    &  1 & 1 \\ \cmidrule{3-6}
                                  &                             & qwen           &  1   & 1 & 1 \\ \cmidrule{3-6}
                                 &                             & deepseek             &  0.3158   & 0.8517 & 0.4615 \\ \hline
\multirow{5}{*}{6}          & \multirow{3}{*}{Standalone LLM} & llama          &  0.0165   &  0.2222 & 0.0308 \\ \cmidrule{3-6}
                                 &                             & qwen           &  0.0069    &  0.3333 & 0.0135 \\ \cmidrule{3-6}
                                 &                             & deepseek             &  0.0179   & 0.2222 & 0.0331 \\ \cmidrule{2-6}
                                 & \multirow{3}{*}{Naive RAG}                   & llama          &  0.0095   & 0.1111 & 0.0175 \\ \cmidrule{3-6}
                                  &                             & qwen           &  0.0085   & 0.1111 & 0.0157 \\ \cmidrule{3-6}
                                 &                             & deepseek             & 0.0357    &  0.2222 & 0.0615 \\ \cmidrule{2-6}
                                 & \multirow{3}{*}{CircuGraphRAG}                    & llama          &   0  & 0 & 0 \\ \cmidrule{3-6}
                                  &                             & qwen           &  1   & 1 & 1 \\ \cmidrule{3-6}
                                 &                             & deepseek             &  0   & 0 & 0 \\ \hline
\end{tabular}
\end{table}

\newpage
\subsection{Single-Hop Timing and Token Usage.}
\label{sec:singlehop-times}
\begin{table}[!ht]
\caption{Single-Hop Timing and Token Usage.}
\label{tab:singlehop-times}
\begin{tabular}{llllll}
\toprule
Case & Method & LLM & Input Tokens & Output Tokens & Inference Time (s) \\
\midrule
\multirow{9}{*}{1} 
& Standalone LLM & llama     & 30   & 106   & 0.39  \\
& Standalone LLM & qwen      & 38   & 10572  & 25.82 \\
& Standalone LLM & deepseek  & 23   & 707   & 3.37  \\
& Naive RAG      & llama     & 396  & 178   & 0.66  \\
& Naive RAG      & qwen      & 555  & 1826   & 4.23  \\
& Naive RAG      & deepseek  & 389  & 699   & 3.35  \\
& CircuGraphRAG  & llama     & 1140 & 66  & 0.27 \\
& CircuGraphRAG  & qwen      & 1220 & 789 & 1.97 \\
& CircuGraphRAG  & deepseek  & 1100 & 1570 & 5.80 \\
\midrule
\multirow{9}{*}{2}
& Standalone LLM & llama     & 28   & 159   & 0.46  \\
& Standalone LLM & qwen      & 30   & 3569  & 8.77 \\
& Standalone LLM & deepseek  & 21   & 995  & 5.34  \\
& Naive RAG      & llama     & 314  & 258   & 0.89  \\
& Naive RAG      & qwen      & 420  & 646   & 1.48  \\
& Naive RAG      & deepseek  & 307  & 681   & 3.03  \\
& CircuGraphRAG  & llama     & 1508 & 71   & 0.30 \\
& CircuGraphRAG  & qwen      & 1206 & 1787 & 4.45 \\
& CircuGraphRAG  & deepseek  & 1192 & 2450 & 9.01 \\
\midrule
\multirow{9}{*}{3}
& Standalone LLM & llama     & 30   & 304   & 0.98  \\
& Standalone LLM & qwen      & 34   & 4075  & 10.01  \\
& Standalone LLM & deepseek  & 23   & 1187  & 4.75  \\
& Naive RAG      & llama     & 319  & 94    & 0.31  \\
& Naive RAG      & qwen      & 426  & 1434   & 4.27  \\
& Naive RAG      & deepseek  & 312  & 1515   & 5.54  \\
& CircuGraphRAG  & llama     & 1297 & 82   & 0.36 \\
& CircuGraphRAG  & qwen      & 1408 & 1355 & 3.44 \\
& CircuGraphRAG  & deepseek  & 1377 & 1692 & 6.24 \\
\bottomrule
\end{tabular}

\vspace{1mm}
\caption*{\footnotesize CircuGraphRAG input, output, and inference time are aggregated across three sequential stages: Template Matching (TM), Query Merging (QM), and Querying with Retrieved Context (QRC). For example, for Case 1–llama: 314 (TM) + 410 (QM) + 416 (QRC) = 1140 input tokens.}
\end{table}


\newpage
\subsection{Multi-Hop Timing and Token Usage.}
\label{sec:multihop-times}
\begin{table}[!ht]
\caption{Multi-Hop Timing and Token Usage.}
\label{tab:multihop-times}
\begin{tabular}{llllll}
\toprule
Case & Method & LLM & Input Tokens & Output Tokens & Inference Time (s) \\
\midrule
\multirow{9}{*}{4}
& Standalone LLM & llama      & 32   & 135    & 0.45  \\
& Standalone LLM & qwen       & 34   & 3764  & 8.58  \\
& Standalone LLM & deepseek   & 25   & 968   & 4.35  \\
& Naive RAG & llama     & 398 & 213  & 0.92 \\
& Naive RAG & qwen      & 547 & 1149 & 2.89 \\
& Naive RAG & deepseek  & 391 & 1251 & 5.19 \\
& CircuGraphRAG  & llama     & 1240 & 76   & 0.36 \\
& CircuGraphRAG  & qwen      & 1322 & 4719 & 11.50 \\
& CircuGraphRAG  & deepseek  & 1153 & 2434 & 8.92 \\
\midrule
\multirow{9}{*}{5}
& Standalone LLM & llama     & 44 & 156  & 0.63 \\
& Standalone LLM & qwen      & 48 & 8677 & 21.35 \\
& Standalone LLM & deepseek  & 37 & 969  & 4.27 \\
& Naive RAG & llama     & 344 & 160  & 0.63 \\
& Naive RAG & qwen      & 449 & 1487 & 3.65 \\
& Naive RAG & deepseek  & 337 & 1043 & 4.43 \\
& CircuGraphRAG  & llama     & 1356 & 79   & 0.29 \\
& CircuGraphRAG  & qwen      & 1506 & 1227 & 3.11 \\
& CircuGraphRAG  & deepseek  & 1409 & 1230 & 4.62 \\
\midrule
\multirow{9}{*}{6}
& Standalone LLM & llama      & 37   & 168  & 0.55  \\
& Standalone LLM & qwen       & 41   & 2172  & 5.30 \\
& Standalone LLM & deepseek   & 30   & 944  & 4.20  \\
& Naive RAG & llama     & 350 & 141  & 0.51 \\
& Naive RAG & qwen      & 441 & 1000 & 2.30 \\
& Naive RAG & deepseek  & 343 & 2230 & 8.13 \\
& CircuGraphRAG  & llama     & 1365 & 90   & 0.35 \\
& CircuGraphRAG  & qwen      & 1390 & 1373 & 3.47 \\
& CircuGraphRAG  & deepseek  & 1321 & 1646 & 6.34 \\
\bottomrule
\end{tabular}
\vspace{1mm}
\caption*{\footnotesize CircuGraphRAG input, output, and inference time are aggregated across three sequential stages: Template Matching (TM), Query Merging (QM), and Querying with Retrieved Context (QRC). For example, for Case 4–llama: 374 (TM) + 468 (QM) + 398 (QRC) = 1240 input tokens.}
\end{table}

\newpage
\subsection{Comparative Analysis of Token Usage and Inference Time Across the Three Stages of CircuGraphRAG (Cases 1--6)}
\label{sec:appendix-circugraphrag}

\renewcommand{\arraystretch}{1.2}
\setlength{\tabcolsep}{6pt}
\scriptsize
\begin{longtable}{ccclll}
\caption{Comparative Analysis of Token Usage and Inference Time Across the Three Stages of CircuGraphRAG (Cases 1--6)}
 \label{tab:appendix-circugraphrag} \\
\toprule
\textbf{Case} & \textbf{LLM} & \textbf{Stage} & \makecell{\textbf{Input}\\\textbf{Tokens}} & \makecell{\textbf{Output}\\\textbf{Tokens}} & \makecell{\textbf{Inference}\\\textbf{Time (s)}} \\
\midrule
\endfirsthead
\multicolumn{6}{c}{{\bfseries Table \thetable\ (continued)}} \\
\toprule
\textbf{Case} & \textbf{LLM} & \textbf{Stage} & \makecell{\textbf{Input}\\\textbf{Tokens}} & \makecell{\textbf{Output}\\\textbf{Tokens}} & \makecell{\textbf{Inference}\\\textbf{Time (s)}} \\
\midrule
\endhead
\midrule \multicolumn{6}{r}{{Continued on next page}} \\
\endfoot
\bottomrule
\endlastfoot

\multirow{9}{*}{\textbf{1}} & \multirow{3}{*}{llama} & Template Matching & 314 & 56 & 0.17 \\
& & Query Merging & 410 & 6 & 0.05 \\
& & Querying with Context & 416 & 4 & 0.05 \\ \cmidrule(lr){2-6}
& \multirow{3}{*}{qwen} & Template Matching & 334 & 481 & 1.13 \\
& & Query Merging & 432 & 165 & 0.46 \\
& & Querying with Context & 454 & 143 & 0.38 \\ \cmidrule(lr){2-6}
& \multirow{3}{*}{deepseek} & Template Matching & 307 & 795 & 2.92 \\
& & Query Merging & 384 & 475 & 1.75 \\
& & Querying with Context & 409 & 300 & 1.13 \\ \midrule

\multirow{9}{*}{\textbf{2}} & \multirow{3}{*}{llama} & Template Matching & 312 & 48 & 0.15 \\
& & Query Merging & 755 & 6 & 0.08 \\
& & Querying with Context & 441 & 17 & 0.07 \\ \cmidrule(lr){2-6}
& \multirow{3}{*}{qwen} & Template Matching & 326 & 1120 & 2.77 \\
& & Query Merging & 412 & 213 & 0.51 \\
& & Querying with Context & 468 & 454 & 1.17 \\ \cmidrule(lr){2-6}
& \multirow{3}{*}{deepseek} & Template Matching & 305 & 1635 & 5.98 \\
& & Query Merging & 372 & 214 & 0.80 \\
& & Querying with Context & 515 & 601 & 2.23 \\ \midrule

\multirow{9}{*}{\textbf{3}} & \multirow{3}{*}{llama} & Template Matching & 338 & 69 & 0.22 \\
& & Query Merging & 423 & 7 & 0.08 \\
& & Querying with Context & 536 & 6 & 0.06 \\ \cmidrule(lr){2-6}
& \multirow{3}{*}{qwen} & Template Matching & 360 & 597 & 1.51 \\
& & Query Merging & 442 & 516 & 1.33 \\
& & Querying with Context & 606 & 242 & 0.60 \\ \cmidrule(lr){2-6}
& \multirow{3}{*}{deepseek} & Template Matching & 331 & 885 & 3.24 \\
& & Query Merging & 392 & 479 & 1.76 \\
& & Querying with Context & 654 & 328 & 1.24 \\ \midrule

\multirow{9}{*}{\textbf{4}} & \multirow{3}{*}{llama} & Template Matching & 374 & 56 & 0.19 \\
& & Query Merging & 468 & 5 & 0.06 \\
& & Querying with Context & 398 & 15 & 0.11 \\ \cmidrule(lr){2-6}
& \multirow{3}{*}{qwen} & Template Matching & 397 & 3691 & 9.05 \\
& & Query Merging & 488 & 745 & 1.74 \\
& & Querying with Context & 437 & 283 & 0.71 \\
& \multirow{3}{*}{deepseek} & Template Matching & 355 & 1275 & 4.66 \\
& & Query Merging & 419 & 741 & 2.72 \\
& & Querying with Context & 379 & 418 & 1.54 \\ \midrule

\multirow{9}{*}{\textbf{5}} & \multirow{3}{*}{llama} & Template Matching & 386 & 56 & 0.17 \\
& & Query Merging & 480 & 6 & 0.05 \\
& & Querying with Context & 490 & 17 & 0.07 \\ \cmidrule(lr){2-6}
& \multirow{3}{*}{qwen} & Template Matching & 411 & 437 & 1.02 \\
& & Query Merging & 500 & 360 & 0.97 \\
& & Querying with Context & 595 & 430 & 1.12 \\ \cmidrule(lr){2-6}
& \multirow{3}{*}{deepseek} & Template Matching & 379 & 483 & 1.79 \\
& & Query Merging & 473 & 385 & 1.42 \\
& & Querying with Context & 557 & 362 & 1.41 \\ \midrule

\multirow{9}{*}{\textbf{6}} & \multirow{3}{*}{llama} & Template Matching & 380 & 75 & 0.23 \\
& & Query Merging & 493 & 13 & 0.07 \\
& & Querying with Context & 492 & 2 & 0.05 \\ \cmidrule(lr){2-6}
& \multirow{3}{*}{qwen} & Template Matching & 405 & 764 & 1.92 \\
& & Query Merging & 476 & 232 & 0.56 \\
& & Querying with Context & 509 & 377 & 0.99 \\ \cmidrule(lr){2-6}
& \multirow{3}{*}{deepseek} & Template Matching & 373 & 491 & 1.83 \\
& & Query Merging & 489 & 561 & 2.10 \\
& & Querying with Context & 459 & 594 & 2.41 \\ 

\end{longtable}

\newpage
\subsection{Response Comparison for General Knowledge Question.}
\label{sec:singlehop}
\begin{longtable}{c|c|p{8cm}}
\caption{Response Comparison for General Knowledge Question\label{tab:singlehop}.} \\
\toprule
\multicolumn{3}{p{12cm}}{\textbf{Question:} What category of solid waste does waste polyurethane belong to?} \\[0.8em]
\multicolumn{3}{p{12cm}}{\textbf{Ground Truth:} {\color{red}\textbf{Construction and demolition wastes}}} \\[0.6em]
\midrule
\textbf{Method} & \textbf{LLM} & \textbf{Response} \\[0.3em]
\midrule
\endfirsthead

\toprule
\textbf{Method} & \textbf{LLM} & \textbf{Response (continued)} \\
\midrule
\endhead

\midrule
\multicolumn{3}{r}{\textit{(continued on next page)}} \\
\endfoot

\bottomrule
\endlastfoot

\multirow{3}{*}[0.5em]{\centering\shortstack{\textbf{Standalone} \\ \textbf{LLM}}}
 & \textbf{Llama} & 
    \begin{minipage}[t]{\linewidth}
    Waste polyurethane (PU) is a type of synthetic polymer waste, and it belongs to the category of non-hazardous solid waste. Specifically, it falls under the category of \textbf{{\color{red}“Municipal Solid Waste” (MSW) or “Residual Waste.”\ding{55}}}

    In the European Union, waste polyurethane is classified under the European Waste Catalogue (EWC) code 17 04 07*, “Polyurethane waste.” This code is part of the “Construction and Demolition Waste” (C\&D waste) sector.

    In the United States, the Environmental Protection Agency (EPA) classifies waste polyurethane as a non-hazardous waste, and it is typically managed through municipal solid waste landfills or recycling facilities.

    It’s worth noting that some polyurethane waste may be considered hazardous if it contains additives or contaminants—such as heavy metals, flame retardants, or volatile organic compounds (VOCs)—that render it hazardous. In such cases, the waste must be managed as hazardous waste under the applicable regulations and protocols.
    \end{minipage} \\
 & \textbf{Qwen}   & 
    \begin{minipage}[t]{\linewidth}
    \textcolor{blue}{\textbf{<think>}}  
    \ldots   
    \textcolor{blue}{\textbf{</think>}} \\[0.1em]

    Waste polyurethane is typically categorized within solid waste management systems based on its source and composition. Here's a structured breakdown:

    \begin{itemize}
        \item \textbf{Municipal Solid Waste (MSW):} When derived from residential or commercial sources (e.g., discarded furniture, packaging, or consumer goods), polyurethane is generally classified as \textbf{municipal solid waste}. It falls under the "plastics" or "textiles/furniture" subcategory in MSW systems, depending on its form (e.g., foam, coatings, or rigid parts).

        \item \textbf{Industrial Waste:} If generated during manufacturing processes (e.g., production scraps or offcuts), it may be categorized as \textbf{industrial waste}, subject to regulations governing industrial byproducts.

        \item \textbf{Construction \& Demolition (C\&D) Debris:} When part of construction materials like insulation, roofing, or building components, it is classified as \textbf{C\&D debris}, a distinct category from household waste.

        \item \textbf{Hazardous Waste:} If the polyurethane contains hazardous additives (e.g., certain flame retardants, isocyanates, or other chemicals that make it toxic, reactive, or flammable), it may qualify as \textbf{hazardous waste} under regulations like the U.S. EPA’s RCRA or the EU’s Waste Framework Directive. This requires specialized handling.

        \item \textbf{Recyclable/Recoverable Waste:} If being diverted for recycling (e.g., into new materials or energy recovery), it may be categorized separately as \textbf{recyclable or recoverable waste}, depending on local recycling infrastructure and policies.
    \end{itemize}

    \textbf{Summary:} \\
    \textbf{Primary Category:} \textbf{{\color{red}Municipal solid waste (non-hazardous) \ding{55}}} when from households or businesses.\\
    \textbf{Exceptions:} \textbf{{\color{red}Hazardous waste \ding{55}}} if contaminated with toxic additives; industrial or C\&D debris depending on origin.

    This classification ensures proper disposal, recycling, or regulation \ldots.  
    \end{minipage} \\
  & \textbf{Deepseek} & 
    \begin{minipage}[t]{\linewidth}
        \textcolor{blue}{\textbf{<think>}}  
    \ldots   
    \textcolor{blue}{\textbf{</think>}} \\[0.1em]
    
    Waste polyurethane is typically categorized as either \textbf{municipal solid waste} or \textbf{industrial solid waste}, depending on its source. 

    \begin{itemize}
        \item \textbf{Municipal Solid Waste}: This applies when polyurethane waste comes from consumer products, such as discarded furniture foam or household items. These items are often sent to landfills as part of general municipal waste.

        \item \textbf{Industrial Solid Waste}: This applies when the waste is generated from manufacturing processes, such as production scraps or leftover materials from industrial applications.
    \end{itemize}

    In cases where polyurethane is contaminated with hazardous substances, it may be classified as \textbf{{\color{red}hazardous waste \ding{55}}}. However, polyurethane itself is generally non-hazardous. Disposal methods often involve landfilling, though recycling is not always feasible due to the material's properties.
    \end{minipage} \\
\midrule

\multirow{3}{*}[1em]{\textbf{Naive RAG}}
  & \textbf{Llama} & 
    \begin{minipage}[t]{\linewidth}
    \textbf{Retrieved Triplets:}\\
    Top results:\\
    1: 3228870b-72a1\ldots-Resource, hasInformation, \\
    'waste polyurethane' is a solid waste under the category of construction and demolition wastes.\ldots . (Distance: 0.4850251078605652)\\[0.2em]
    \textit{(Note: All LLMs use the same retrieved context information.)}\\[2em]
    Based on the retrieved contexts, the answer to the question is that \textbf{\color{red}waste polyurethane belongs to the category of construction and demolition wastes \ding{51}}. It can also belong to the following waste categories: \textbf{industrial}.
    \end{minipage} \\
\cmidrule(lr){2-3}
  & \textbf{Qwen} & \begin{minipage}[t]{\linewidth}

      \textcolor{blue}{\textbf{<think>}}
      Okay, let's see. The user is asking which category of solid waste waste polyurethane belongs to. \ldots The answer is that it's under construction and demolition wastes, and can also be industrial. \textcolor{blue}{\textbf{</think>}}\\
      \\[0.1em]
        The waste polyurethane is categorized as a \textbf{\color{red}solid waste under the category of construction and demolition wastes \ding{51}}. Additionally, depending on its source, it may also fall under the \textbf{industrial waste category}. This classification is based on its generation during production and construction activities across various sectors, as stated consistently across all retrieved contexts.
    \end{minipage} \\
  & \textbf{Deepseek} & \begin{minipage}[t]{\linewidth}
      \textcolor{blue}{\textbf{<think>}}
      Okay, I need to figure out what category of solid waste waste polyurethane belongs to. \ldots  It’s non-hazardous and has specific disposal methods like landfills and incineration. The contexts are consistent across all five, so I can confidently state the category based on this information. \textcolor{blue}{\textbf{</think>}}\\
      \\[0.1em]
        Waste polyurethane belongs to the category of \textbf{\color{red}construction and demolition wastes \ding{51}} and can also be classified under industrial waste categories. It is considered non-hazardous and is typically disposed of through methods such as landfill deposition or municipal incineration.
    \end{minipage} \\
\midrule

\multirow{3}{*}[1em]{\textbf{CircuGraphRAG}}
  & \textbf{Llama}    & \begin{minipage}[t]{\linewidth}
      \textbf{1. Template Matching:}\\[0.3em]
      [\{"query": 17, "input": "waste polyurethane", "output": "category of solid waste"\}, \{"query": 6, "input": "waste polyurethane", "output": "resource code"\}]\\

      \textbf{2. Query Merging:}\\[0.5em]
      17\\

      \textbf{3. Querying with Retrieved Context:}\\[0.3em]
      \textbf{\color{red}construction and demolition wastes, industrial \ding{51}}
    \end{minipage} \\
\cmidrule(lr){2-3}
  & \textbf{Qwen} & \begin{minipage}[t]{\linewidth}
      \textbf{1. Template Matching:}\\[0.3em]
      \quad \textcolor{blue}{\textbf{<think>}} \ldots \textcolor{blue}{\textbf{</think>}}\\[0.3em]
      - Possible Solutions: [\{"query": 17, "input": "waste polyurethane", "output": "category information"\}]\\

      \textbf{2. Query Merging:}\\[0.5em]
      \quad \textcolor{blue}{\textbf{<think>}} \ldots \textcolor{blue}{\textbf{</think>}}\\[0.3em]
      17\\

      \textbf{3. Querying with Retrieved Context:}\\[0.3em]
      \quad \textcolor{blue}{\textbf{<think>}} \ldots \textcolor{blue}{\textbf{</think>}}\\[0.3em]
      \textbf{\color{red}construction and demolition wastes \ding{51}}
    \end{minipage} \\
  & \textbf{Deepseek} &  \begin{minipage}[t]{\linewidth}
      \textbf{1. Template Matching:}\\[0.3em]
      \quad \textcolor{blue}{\textbf{<think>}} \ldots \textcolor{blue}{\textbf{</think>}}\\[0.3em]
      [\{"query": 17, "input": "waste polyurethane", "output": "category of solid waste"\}]\\

      \textbf{2. Query Merging:}\\[0.5em]
      \quad \textcolor{blue}{\textbf{<think>}} \ldots \textcolor{blue}{\textbf{</think>}}\\[0.3em]
      Since the question directly asks for the category of solid waste for "waste polyurethane," and query 17 is specifically designed to retrieve information based on the entity's name, the solution is to execute query 17 with the input "waste polyurethane."\\[0.2em]
      \textbf{Merging method}: 17("waste polyurethane")\\

      \textbf{3. Querying with Retrieved Context:}\\[0.3em]
      \quad \textcolor{blue}{\textbf{<think>}} \ldots \textcolor{blue}{\textbf{</think>}}\\[0.3em]
      The category of solid waste that waste polyurethane belongs to is \textbf{\color{red}construction and demolition wastes \ding{51}}, and it can also be classified as industrial waste.
    \end{minipage} \\
\end{longtable}
\noindent\textit{Note: For brevity, the \texttt{<think>} sections are partially omitted in this table.}

\newpage
\subsection{Response Comparison for Multi-hop Question.}
\label{sec:multihop}
\begin{longtable}{c|c|p{7cm}}
\caption{Response Comparison for Multi-hop Question\label{tab:multihop}.} \\
\toprule
\multicolumn{3}{p{12cm}}{\textbf{Question:} The resources generated by the providers under the NACE code 3821 are all categorized under what CPA code?} \\[0.8em]
\multicolumn{3}{p{12cm}}{\textbf{Ground Truth:} {\color{red}\textbf{382150}}} \\[0.6em]
\midrule
\textbf{Method} & \textbf{LLM} & \textbf{Response} \\[0.3em]
\midrule
\endfirsthead

\toprule
\textbf{Method} & \textbf{LLM} & \textbf{Response (continued)} \\
\midrule
\endhead

\midrule
\multicolumn{3}{r}{\textit{(continued on next page)}} \\
\endfoot

\bottomrule
\endlastfoot

\multirow{3}{*}[1em]{\textbf{Standalone LLM}} 
  & \textbf{Llama} &
\begin{minipage}[t]{\linewidth}
According to the European Commission's website, the NACE code 3821 corresponds to \textbf{``Renting of air transport equipment''} and is categorized under CPA (Common Procurement Vocabulary) code {\color{red}\textbf{63500000 \ding{55}}}, which covers \textbf{``Air transport services''}.\\[0.3em]
\textbf{Breakdown:}
\begin{itemize}
    \item \textbf{NACE code 3821:} Renting of air transport equipment
    \item \textbf{CPA code 63500000:} Air transport services
\end{itemize}
\end{minipage}\\
\cmidrule(lr){2-3}
  & \textbf{Qwen}   & 
    \begin{minipage}[t]{\linewidth}
      \textcolor{blue}{\textbf{<think>}} Okay, let me try to figure out the answer to this question. \ldots Therefore, the resources (products) from NACE 3821 would be under CPA code 3111. \textcolor{blue}{\textbf{</think>}}\\
      \\[0.1em]
      The resources generated by providers under the NACE code \textbf{38.21} ("Manufacture of semiconductors and discrete components") are categorized under \textbf{CPA code 3111} in the Classification of Products by Activity (CPA).\\[0.3em]
      \textbf{Breakdown:}\\
      \quad - \textbf{NACE 38.21} → Production of semiconductors and discrete electronic components.\\
      \quad - \textbf{CPA 31} → "Manufacture of computer, electronic and optical products".\\
      \quad - \textbf{CPA 3111} → "Semiconductors and discrete components".\\[0.3em]
      {\color{red}\textbf{Thus, the CPA code for these resources is 3111 \ding{55}}}. This ensures consistency between the economic activity and its product classification.\\
      \textbf{Note:} Always verify with the latest CPA version, as codes may be updated.
    \end{minipage} \\
  & \textbf{Deepseek} & 
  \begin{minipage}[t]{\linewidth}
  \textcolor{blue}{\textbf{<think>}}Okay, so I need to figure out the CPA code that categorizes the resources generated by providers under the NACE code 3821. I'm not entirely sure what NACE and CPA codes are, but I think they're some kind of classification systems used in the EU. \ldots Yes, I think that's correct. So, the resources (services) from NACE 3821 are categorized under CPA code 37.10.00. \textcolor{blue}{\textbf{</think>}}\\
  \\[0.1em]
  The resources generated by providers under the NACE code 3821, which pertains to waste collection (including both hazardous and non-hazardous waste), are categorized under the CPA code 37.10.00. This CPA code specifically covers waste collection services.\\[0.3em]
\textbf{Answer:} {\color{red}\textbf{CPA code 37.10.00 \ding{55}}}
\end{minipage}\\
\midrule

\multirow{3}{*}[1em]{\textbf{Naive RAG}}
  & \textbf{Llama}    & \begin{minipage}[t]{\linewidth}
      \textbf{Retrieved Triplets:}\\
      \texttt{1: 37f20209-b2a3-4950-ac0d-88c830c73c\\
      2c, hasResource,}\\
      \texttt{37f20209-b2a3-4950-ac0d-88c830c73c2c\\
      \_Resource (Distance: 1.03)}\\
      \texttt{\ldots}\\[0.2em]
      \textit{(Note: All LLMs use the same retrieved context information.)}\\[2em]
     \textbf{Response:}\\[0.1em]
     Based on the provided context, there is no direct information about NACE codes or CPA codes. The context only mentions NLP derived EWC, Translated CPA, and NLP derived HScode, but it does not provide any specific values or mappings between them. Therefore, I'll have to rely on my own knowledge.\\[0.5em]
     NACE (Nomenclature of Economic Activities) is a European classification system for industries and economic activities. CPA (Classification of Products by Activity) is a classification system for products. {\color{red}\textbf{Unfortunately, without more specific information, it's not possible to determine the exact CPA code corresponding to NACE code 3821 \ding{55}}}. NACE code 3821 corresponds to "Manufacture of bicycles and invalid carriages", but the CPA codes for products related to this industry can vary depending on the specific product. \\[0.3em]
     If you have more information or context about the resources generated by the providers, I might be able to help you with the CPA code categorization.
    \end{minipage} \\
  & \textbf{Qwen} & \begin{minipage}[t]{\linewidth}
      \textcolor{blue}{\textbf{<think>}} Okay, let's see. The user is asking about the CPA code that categorizes resources from providers with the NACE code 3821. \ldots Since the user is expecting an answer, maybe the intended answer is CPA 24. I'll go with that, but note that the given contexts don't explicitly state it. \textcolor{blue}{\textbf{</think>}}\\
      \\[0.1em]
      The resources retrieved do not explicitly mention NACE code 3821 or any CPA code in their triplet structures. However, based on standard classification mappings between NACE and CPA codes, \textbf{NACE code 38.21 ("Manufacture of office, accounting, and computing machinery")} typically corresponds to CPA code 24.11 ("Computing machinery and software").\\[0.3em]
      If forced to infer from the closest context (the first triplet, with the shortest distance of 1.0286), the answer might align with common categorizations, but this is speculative. The provided contexts lack explicit links between NACE 3821 and CPA codes, so the answer is best approximated using external knowledge of standard classifications.\\[0.3em]
      \textbf{Answer:} The resources under NACE code 3821 are generally categorized under {\color{red}\textbf{CPA code 24.11 \ding{55}}}. However, this requires confirmation from the actual resource data linked in the graph database.
    \end{minipage} \\
\cmidrule(lr){2-3}
  & \textbf{Deepseek} & \begin{minipage}[t]{\linewidth}
      \textcolor{blue}{\textbf{<think>}} Okay, I'm trying to figure out the answer to the question: "The resources generated by the providers under the NACE code 3821 are all categorized under what CPA code?". Let me break this down step by step. First, I know that NACE codes are used for classifying businesses in the European Union. Specifically, NACE code 3821 refers to "Waste treatment and disposal". So, the question is asking about resources generated by providers in this waste treatment sector. \ldots However, without specific CPA code numbers from the context, I can't provide an exact code. The context only mentions that CPA codes are translated from NLP outputs, but doesn't specify which ones. Therefore, based on my own knowledge, I can infer that the CPA codes for resources under NACE 3821 would be within the waste treatment and disposal category, possibly starting with 38. But I can't provide the exact code without more specific information. \textcolor{blue}{\textbf{</think>}}\\
      \\[0.1em]
      \textbf{Response:}\\[0.1em]
      The resources generated by providers under the NACE code 3821, which pertains to waste treatment and disposal, are categorized under CPA codes related to waste management. Specifically, these resources fall under  {\color{red}\textbf{CPA code 38.00.00 \ding{55}}}, which covers "Waste treatment and disposal."
    \end{minipage} \\

\multirow{3}{*}{\textbf{CircuGraphRAG}}
  & \textbf{Llama}    &  \begin{minipage}[t]{\linewidth}
    \textbf{1. Template Matching:}\\[0.3em]
       [\{"query": 10, "input": "NACE code 3821", "output": "resource name"\}, \{"query": 5, "input": "resource name", "output": "CPA code"\}]\\[0.1em] 

      \textbf{2. Query Merging:}\\[0.5em]
      \quad \textbf{Merging method:} Query 10 $\Rightarrow$ Query 5 \\[0.1em]

      \textbf{3. Querying with Retrieved Context:}\\[0.3em]
      The SPARQL query returned no results. \\[0.1em]
      The providers under NACE code 3821 are likely related to waste treatment, so the resources generated might be categorized under  {\color{red}\textbf{CPA code 38 (Waste collection, treatment and disposal activities; materials recovery) \ding{55}}}.
    \end{minipage} \\
\cmidrule(lr){2-3}
  & \textbf{Qwen} & \begin{minipage}[t]{\linewidth}
      \textbf{1. Template Matching:}\\[0.3em]
      \quad \textcolor{blue}{\textbf{<think>}} \ldots \textcolor{blue}{\textbf{</think>}}\\[0.3em]
      \quad • Query 1: Input = NACE 3821 → Output = Providers under that code\\
      \quad • Query 7: Input = Provider names → Output = Resources generated by providers\\
      \quad • Query 6: Input = Resource names → Output = CPA codes (resource codes)\\[0.1em]

      \textbf{2. Query Merging:}\\[0.5em]
      \quad \textcolor{blue}{\textbf{<think>}} \ldots \textcolor{blue}{\textbf{</think>}}\\[0.3em]
      \quad \textbf{Merging Strategy:} Chaining\\
      \quad \textbf{Execution Flow:} Query 1 $\Rightarrow$ Query 7 $\Rightarrow$ Query 6\\[0.1em]

      \textbf{3. Querying with Retrieved Context:}\\[0.3em]
      \quad \textcolor{blue}{\textbf{<think>}} \ldots \textcolor{blue}{\textbf{</think>}}\\[0.3em]
      \quad The SPARQL query returned the code cpa: 382150, hence the CPA code that categorizes the resources generated by providers under NACE 3821 is {\color{red}\textbf{382150 \ding{51}}}.
    \end{minipage} \\
  & \textbf{Deepseek} & \begin{minipage}[t]{\linewidth}
  \textbf{1. Template Matching:}\\[0.3em]
      \quad \textcolor{blue}{\textbf{<think>}} \ldots \textcolor{blue}{\textbf{</think>}}\\[0.3em]
       - Possible Solutions: [\{"query": 18, "input": "NACE code 3821", "output": "provider information (including resources)"\}, \{"query": 5, "input": "resource code(s) from provider", "output": "CPA code"\}]\\[0.1em] 

      \textbf{2. Query Merging:}\\[0.5em]
      \quad \textcolor{blue}{\textbf{<think>}} \ldots \textcolor{blue}{\textbf{</think>}}\\[0.3em]
      The merging method is chaining because the output of query 18 (resource codes) is used as the input for query 5 to get the CPA codes. \\
      \quad \textbf{Merging method:} Query 18 $\Rightarrow$ Query 5 \\[0.1em]

      \textbf{3. Querying with Retrieved Context:}\\[0.3em]
      \quad \textcolor{blue}{\textbf{<think>}} \ldots \textcolor{blue}{\textbf{</think>}}\\[0.3em]
      The SPARQL query returned no results. Based on general knowledge, resources under NACE code 3821 (waste collection, treatment, and disposal activities) are typically categorized under CPA code 37.20 (Waste collection, treatment, and disposal activities). \\[0.8em]
      Answer: {\color{red}\textbf{CPA code 37.20 \ding{55}}}.
    \end{minipage} \\
\end{longtable}
\noindent\textit{Note: For brevity, the \texttt{<think>} sections are partially omitted in this table.}

\newpage
\subsection{List of Abbreviations.}
\label{sec:Abbreviations}
\renewcommand{\arraystretch}{1.7} 
\begin{longtable}{p{2.3cm} p{4.5cm} p{5cm}}
\caption{List of Abbreviations.} \label{tab:extended-abbrev} \\
\toprule
\textbf{Abbreviation} & \textbf{Full Name} & \textbf{Explanation} \\
\midrule
\endfirsthead

\toprule
\textbf{Abbreviation} & \textbf{Full Name} & \textbf{Explanation} \\
\midrule
\endhead

\midrule
\multicolumn{3}{r}{\textit{(continued on next page)}} \\
\endfoot

\bottomrule
\endlastfoot

LLM & Large Language Model & AI model trained on large corpora for text generation. \\
LCS & Longest Common Subsequence & The longest ordered subsequence shared by two sequences, allowing gaps but preserving order. \\
RAG & Retrieval-Augmented Generation & Combines an LLM with external data or knowledge bases to reduce hallucinations. \\
CircuGraphRAG & Circular Graph-based RAG & Proposed system anchoring the LLM in a waste-to-resource knowledge graph. \\
GraphRAG & Graph-based RAG & RAG that uses knowledge graphs for multi-hop queries. \\
ESG & Environmental, Social, and Governance & Criteria for evaluating organizations’ sustainability and societal impact. \\
EWC & European Waste Catalogue & EU classification system for listing and categorizing waste. \\
NACE & Nomenclature of Economic Activities & EU classification of economic activities. \\
CPA & Classification of Products by Activity & Connects products to the economic activity that produces them. \\
HS & Harmonized System & Globally standardized codes for classifying traded goods. \\
ISIC & International Standard Industrial Classification & UN classification of global economic activities. \\
WZ & Wirtschaftszweige & German classification of economic activities. \\
SSIC & Singapore Standard Industrial Classification & Singapore’s national standard for economic activities. \\
TCFD & Task Force on Climate-related Financial Disclosures & Framework for climate-related financial transparency. \\
GRI & Global Reporting Initiative & Guidelines for sustainability and social responsibility reporting. \\
SASB & Sustainability Accounting Standards Board & Industry-specific sustainability disclosure standards. \\
RDF & Resource Description Framework & W3C standard for implementing knowledge graphs. \\
GWP100 & 100-year Global Warming Potential & A gas’s heat-trapping impact over 100 years vs.\ CO\(_2\). \\
PU & Polyurethane & Synthetic polymer used in foams and coatings. \\
C\&D & Construction and Demolition & Waste generated by construction/demolition activities. \\
EPA & Environmental Protection Agency & U.S. environmental regulatory authority. \\
VOC & Volatile Organic Compound & Quickly evaporating organic pollutants. \\
RCRA & Resource Conservation and Recovery Act & U.S. law on solid and hazardous waste disposal. \\
SPARQL & SPARQL Protocol and RDF Query Language & Query language for RDF-based knowledge graphs. \\
NLP & Natural Language Processing & AI for analyzing and generating human language. \\
Ecoinvent & \textit{Ecoinvent Database} & Industrial life cycle database with resource and emissions data. \\
FAISS & Facebook AI Similarity Search & Library for vector similarity search and clustering. \\
RDFlib & RDF Library & Python library for RDF graph handling. \\
SPARQLWrapper & SPARQLWrapper Library & Python tool for sending SPARQL queries to endpoints. \\
Groq (API) & Groq API & High-performance runtime for LLM inference. \\
MPNet & Masked and Permuted Pre-training & Transformer model for sentence embeddings. \\

\end{longtable}